\pdfoutput=1 % arXiv: force pdfLaTeX; must stay within the first 5 lines
\documentclass{article}
\usepackage{arxiv}

\usepackage[utf8]{inputenc}
\usepackage[T1]{fontenc}
\usepackage{microtype}
\usepackage{graphicx}
\usepackage{natbib}
\setcitestyle{authoryear,round,citesep={;},aysep={,},yysep={;}}

\usepackage{amsmath,amsfonts,bm}

\def\eqref#1{equation~\ref{#1}}
\def\1{\bm{1}}

\DeclareMathAlphabet{\mathsfit}{\encodingdefault}{\sfdefault}{m}{sl}
\SetMathAlphabet{\mathsfit}{bold}{\encodingdefault}{\sfdefault}{bx}{n}

\usepackage{hyperref}
\usepackage{url}
\usepackage{booktabs,tabularx,flafter}
\usepackage{pifont}
\usepackage[table]{xcolor}

\usepackage{xspace}
\usepackage{flushend}
\usepackage{enumitem}
\usepackage[ruled,vlined,linesnumbered]{algorithm2e}
\usepackage{multicol}
\usepackage{xcolor}
\usepackage[skins]{tcolorbox}
\usepackage{placeins}
\usepackage{amsmath}
\usepackage{bm}
\usepackage{subcaption}
\usepackage{amsthm}
\usepackage{listings}
\theoremstyle{acmdefinition}

\usepackage{multirow}
\usepackage{booktabs}
\usepackage{tabularx}
\usepackage{pifont}
\usepackage[normalem]{ulem}
\usepackage{color, colortbl}
\usepackage{makecell}
\usepackage{dashbox}
\usepackage{ragged2e}

\title{\tool: Teaching Visual Game Agents to Play with Programmatic Experts as Demonstrators and Step-Level Judges}

\author{%
\begin{minipage}{0.95\textwidth}
\centering\normalfont
Zhuotong Zhou\textsuperscript{1} \quad
Qian Zeng\textsuperscript{2} \quad
Zijun Liu\textsuperscript{3} \quad
Wentao Liu\textsuperscript{4} \\
Biao Yuan\textsuperscript{5} \quad
Siyu Song\textsuperscript{6} \quad
Jingying Zhou\textsuperscript{7} \quad
Youjia Li\textsuperscript{3} \\
Xiaokun Wang\textsuperscript{8,\dag} \quad
Min Ji\textsuperscript{3,\dag} \quad
Ji Wang\textsuperscript{9,\dag} \\[1.5em]
\textsuperscript{1}Fudan University \quad
\textsuperscript{2}Zhejiang University \\
\textsuperscript{3}Shanghai Tianyou Software Co., Ltd. \quad
\textsuperscript{4}Shanghai Institute of Innovation \\
\textsuperscript{5}Anhui University \quad
\textsuperscript{6}East China Normal University \quad
\textsuperscript{7}University of Melbourne \\
\textsuperscript{8}Chabiyue (Shanghai) Information Technology Co., Ltd. \\
\textsuperscript{9}Zhejiang Century Huatong Group Co., Ltd. \\[1.5em]
\textsuperscript{\dag}Corresponding authors. \\
Xiaokun Wang: \texttt{chubbyue@163.com}; \quad
Min Ji: \texttt{M.JI@t2cn.com} \\
Ji Wang: \texttt{vipwj@iCloud.com}
\end{minipage}%
}

\renewcommand{\shorttitle}{\tool: Teaching Visual Game Agents with Programmatic Experts}

\hypersetup{
  colorlinks=true,
  linkcolor=red!60!black,
  citecolor=blue!60!black,
  urlcolor=magenta!70!black,
  pdftitle={Apprentice: Teaching Visual Game Agents to Play with Programmatic Experts as Demonstrators and Step-Level Judges},
  pdfauthor={Zhuotong Zhou, Qian Zeng, Zijun Liu, Wentao Liu, Biao Yuan, Siyu Song, Jingying Zhou, Youjia Li, Xiaokun Wang, Min Ji, Ji Wang},
  pdfsubject={cs.AI},
  pdfkeywords={game agents, gameplay trajectories, vision-language models},
}

\newcolumntype{C}[1]{>{\centering\arraybackslash}p{#1}}

\newcommand{\tool}{\textsc{Apprentice}\xspace}
\newcommand{\todo}[1]{#1}
\definecolor{revcolor}{HTML}{00A3CC}
\newcommand{\rev}[1]{{\color{revcolor}#1}}
\newenvironment{revblock}{\color{revcolor}}{}
\renewcommand{\rev}[1]{#1}\renewenvironment{revblock}{}{}

\begin{document}

\maketitle

\begin{abstract}
% !TeX root = ../main.tex

\rev{Training visual game agents that map multimodal observations directly to native keyboard-and-mouse actions is challenging: offline imitation suffers from costly data collection, diluted action supervision, and poor generalization to unseen states, whereas online reinforcement learning struggles with sparse terminal rewards over long horizons.
We propose \tool, a three-stage framework that trains a unified visual game agent across diverse games, leveraging lightweight programmatic experts that read internal game states during training as both offline demonstrators and online step-level judges.
First, programmatic experts synthesize verified gameplay trajectories at 
high success rates, which are annotated offline with step-level rationales 
and goals.
Second, Action-Focused Supervised Fine-Tuning cold-starts the agent while counteracting supervision dilution from verbose rationales and repetitive waiting steps without extra inference latency.
Third, Expert-Guided On-Policy Learning updates the agent over truncated rollouts via criticality-adaptive on-policy imitation and expert-rewarded policy optimization, providing dense step-level supervision to generalize to unseen states.
Across \todo{45} tasks from \todo{nine} games, our trained agent outperforms the strongest VLM baseline by \todo{39.7\%} in success rate, \todo{38.6\%} in progress rate, and \todo{40.0\%} in task coverage.}

\end{abstract}

% \keywords{Game agents \and Gameplay trajectories \and Vision-language models}

% !TeX root = ../main.tex

\section{Introduction}\label{introduction}

\captionsetup[subfigure]{font=small, skip=-5pt, justification=centering}
\captionsetup{justification=centering, singlelinecheck=false} % 主标题
\begin{figure}[!t]
    \centering
    \includegraphics[width=\linewidth]{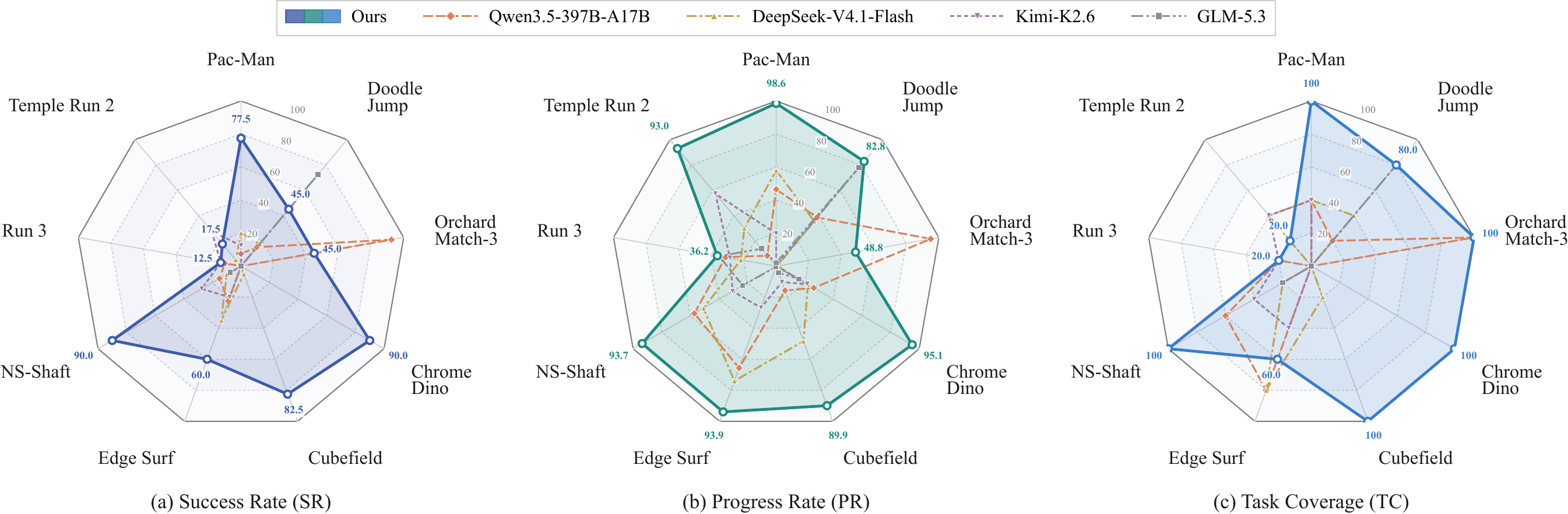}
    \vspace{-1.2em}
    \caption{\justifying Per-game comparison with four strong VLM baselines on (a) success rate, (b) progress rate, and (c) task coverage. Built on Qwen3.5-4B, \tool substantially outperforms much larger VLMs across 45 tasks in nine browser games.}
    \label{fig:teaser}
    \vspace{-1em}
\end{figure}

\looseness=-1 Training embodied agents that perceive multimodal observations and execute actions in 
dynamic environments is a central pursuit in artificial intelligence~\citep{brohan2023rt2,
kim2024openvla,black2024pi0,sima2025}.
Video games provide an ideal, controllable testbed for this direction, combining rich visual dynamics and general-purpose input interfaces with verifiable task outcomes~\citep{zhang2025videogamebench,ouyang2026gameworld}.
However, a visual game agent that relies solely on multimodal observations and native keyboard-and-mouse controls faces two fundamental challenges.
First, it must rapidly ground these observations into accurate low-level actions. Without access to internal game states, the agent has to infer spatial layouts and object motion directly from screenshots and device states.
Second, gameplay demands decision-dense, long-horizon control over hundreds of steps. While most steps involve passive waiting, a single mistimed action at a critical moment can cause compounding errors or immediate failure.

Existing approaches struggle to address both challenges simultaneously.
Text-based agents rely on symbolic state encodings and game-specific APIs~\citep{wang2023voyager,zhu2023gitm}, which simplifies high-level planning but bypasses visual perception and native device control.
GUI-based agents move closer to human gameplay by perceiving screenshots and acting via keyboard and mouse~\citep{tan2024cradle,wang2025gametars,tan2025lumine}, yet they often still rely on heavy game-specific harnesses.
Meanwhile, training end-to-end visual game agents via imitation learning~\citep{baker2022vpt,sima2024,li2025jarvisvla,magne2026nitrogen} or reinforcement learning (RL)~\citep{shi2026odysseus} faces coupled bottlenecks on decision-dense trajectories.
Offline imitation requires abundant successful trajectories that current vision-language models (VLMs) fail to collect efficiently, while repetitive waiting steps and verbose reasoning traces dilute supervision on concise actions.
Moreover, during inference, the agent inevitably strays into unseen game states, 
suffering from exposure bias~\citep{ross2011dagger}.
Online RL trains on the agent's own rollouts, yet sparse terminal rewards hinder step-level credit assignment over long horizons.

To overcome these limitations, we propose \tool, a three-stage framework that trains a single visual game agent across diverse games within a shared observation-to-action space.
Our key insight is that during training, game environments expose structured internal states from which lightweight, heuristic \emph{programmatic experts} can determine reliable actions, serving first as offline demonstrators and then as online step-level judges.
First, these experts synthesize verified trajectories at high success rates, which are annotated offline with step-level rationales and goals.
Second, we cold-start the agent via \emph{Action-Focused Supervised Fine-Tuning (SFT)}: action-first prediction outputs the action before the rationale and goal and decodes only the action at inference, while intra-step field rebalancing and temporal step reweighting prevent verbose rationales and repetitive waiting steps from diluting action supervision.
Third, to mitigate exposure bias without relying on sparse terminal rewards, we perform \emph{expert-guided on-policy learning} over truncated rollouts, querying the programmatic expert at each visited state for dense step-level supervision.
Specifically, we introduce a simulation-based \emph{step-criticality measure} to quantify each state's error tolerance, and use it to reweight both \emph{on-policy imitation}, which aligns visited states with expert actions, and \emph{Expert-Rewarded Policy Optimization (ERPO)}, which scores sampled candidate actions against the expert for group-relative updates.

We evaluate \tool across \todo{nine} browser games spanning four genres and \todo{45} tasks of increasing difficulty.
Built on Qwen3.5-4B~\citep{qwen3.5}, \tool trains a single visual game agent that achieves an overall success rate of \todo{57.8\%}, a progress rate of \todo{81.3\%}, and a task coverage of \todo{75.6\%}, outperforming the strongest open-source VLM baseline by \todo{39.7\%}, \todo{38.6\%}, and \todo{40.0\%}, respectively (Fig.~\ref{fig:teaser}).
Extensive ablations further confirm that each component of \tool contributes substantially to final gameplay performance.

Our main contributions are as follows:
\begin{itemize}[leftmargin=*,itemsep=1pt,topsep=2pt]
    \item We propose \tool, a three-stage framework for training a unified visual game agent with programmatic experts as offline demonstrators and online step-level judges.
    \item We introduce action-focused SFT to counteract supervision dilution without inference overhead, followed by criticality-adaptive on-policy imitation and expert-rewarded policy optimization over truncated rollouts to generalize to game states unseen during SFT.
    \item Extensive experiments on \todo{45} tasks across \todo{nine} games show that our trained agent attains a \todo{57.8\%} overall success rate, substantially outperforming larger VLM baselines, while thorough ablations validate each component.
\end{itemize}

% !TeX root = ../main.tex

\section{Related Work}\label{related_work}
\textbf{LM-based Game Agents.} Existing LM-based game agents can be broadly divided into two categories: text-based agents and GUI-based agents.
% These categories differ primarily in their observation and action spaces. 
In the first category, agents receive game states symbolically encoded as text through hand-designed interfaces and act through game-specific APIs. For example, Voyager~\citep{wang2023voyager} uses an automatic exploration curriculum to continually acquire new executable skills for playing Minecraft.
% GITM~\citep{zhu2023gitm} further leverages textual knowledge and memory to decompose high-level goals into structured action plans. 
% \rev{Although textual game states and predefined game commands simplify long-horizon planning, they spare these agents from perceiving the game visually and operating the keyboard and mouse, as human players do.}
In the second category, agents perceive games through multimodal observations that combine gameplay screenshots with text-encoded game states and interact through native keyboard and mouse actions. Cradle~\citep{tan2024cradle} proposes a modular agent framework that coordinates planning, reflection, and memory for screenshot-based game interaction. Game-TARS~\citep{wang2025gametars} learns native device actions together with sparse reasoning, while Lumine~\citep{tan2025lumine} learns gameplay control, instruction following, and reasoning capabilities from annotated trajectories. 
However, these methods still rely on heavy game-specific harnesses. \rev{Following the second category, \tool instead trains a single visual game agent that maps screenshots and device states directly to native keyboard and mouse controls, establishing a shared observation and action space across games.}

\textbf{\rev{Imitation Learning.}} \rev{VLA models learn to map visual and textual inputs to executable action sequences by imitating demonstrations.}
% In robotics, RT-2~\citep{brohan2023rt2} predicts robot actions as text tokens, OpenVLA~\citep{kim2024openvla} learns arm and gripper control from recorded robot trajectories, and $\pi_0$~\citep{black2024pi0} generates commands for multiple future steps through flow matching.
In games, SIMA~\citep{sima2024} and SIMA~2~\citep{sima2025} learn keyboard and mouse control across virtual worlds, and JARVIS-VLA~\citep{li2025jarvisvla} trains a VLM on Minecraft gameplay trajectories.
VPT~\citep{baker2022vpt}, MineRL~\citep{guss2019minerl}, NitroGen~\citep{magne2026nitrogen}, and Open P2P~\citep{yue2026p2p} further scale imitation to large collections of human gameplay and online videos. 
However, behavioral cloning on fixed demonstrations suffers from exposure bias, causing compounding errors in unseen states.
DAgger~\citep{ross2011dagger} mitigates this issue by iteratively querying an expert on states visited by the learner and aggregating these labeled rollouts into the training set. Going beyond DAgger, we design programmatic experts to guide a progressive training pipeline: offline cold-start training followed by online expert imitation.

\textbf{Game Benchmarks.} \rev{Game benchmarks~\citep{huang2024gamabench,costarelli2024gamebench,wu2023smartplay,wang2025lvlmplayground,paglieri2025balrog,zheng2026vmage,hu2026lmgame,park2026orak} evaluate LMs in gameplay, ranging from strategic reasoning to visual perception and control.}
% GAMA-Bench~\citep{huang2024gamabench}, GameBench~\citep{costarelli2024gamebench}, and SmartPlay~\citep{wu2023smartplay} assess planning, decision-making, and reasoning through structured textual interactions \rev{without visual perception}.
% \rev{LVLM-Playground~\citep{wang2025lvlmplayground} and BALROG~\citep{paglieri2025balrog} incorporate visual board and grid environments, and V-MAGE~\citep{zheng2026vmage}, LMGame-Bench~\citep{hu2026lmgame}, and Orak~\citep{park2026orak} extend evaluation to more complex video games.}
However, most of them constrain agents to textual state representations and game-specific control APIs. In contrast, VideoGameBench~\citep{zhang2025videogamebench} and GameWorld~\citep{ouyang2026gameworld} support multimodal perception and native keyboard and mouse interaction. We adopt GameWorld as our environment.

% \rev{\textbf{Reinforcement Learning for VLM Agents.}
% Policy-gradient methods such as PPO~\citep{schulman2017ppo} and GRPO~\citep{shao2024deepseekmath} have become standard for fine-tuning language and vision-language models with verifiable rewards.
% Odysseus~\citep{shi2026odysseus} applies RL to VLMs in games with more than a hundred decision turns.
% Such methods typically reward an episode or a response by its final outcome, which becomes sparse and delayed in real-time games with hundreds of control steps.
% Instead, we obtain a dense reward at every step from the judgment of a programmatic expert on the state visited by the agent.}

% !TeX root = ../main.tex

\section{\texorpdfstring{\rev{Preliminaries}}{Preliminaries}}\label{sec:preliminaries}\label{sec:visual-game-agent}
We study visual game agents that interact with game environments via multimodal observations and device actions, and formalize this setting below.

\rev{\paragraph{Problem formulation.}
Each game comprises multiple tasks of varying difficulty. Across all tasks, gameplay is inherently partially observable (e.g., a single screenshot does not capture object motion or off-screen objects). 
We therefore model each task as a partially observable Markov decision process (POMDP) $\langle D, \mathcal{Z}, \mathcal{A}, \mathcal{O}, \mathcal{T} \rangle$. Here, $D$ is a natural-language context specifying the game rules, legal controls, and task objective; $\mathcal{Z}$ is the space of latent game states; $\mathcal{A}$ is the space of device actions, detailed below; $\mathcal{O}$ is the space of multimodal observations, where each observation $o = (v, s)$ consists of a screenshot $v$ and the device state $s$ (held keys, held mouse buttons, and cursor position); and the state transition function $\mathcal{T}: \mathcal{Z} \times \mathcal{A} \rightarrow \mathcal{Z}$ models the evolution of the game state in response to actions.

The environment pauses every $50\,\mathrm{ms}$ of game time, each pause marking a decision step. At step $t$, the agent receives the context $D$, the multimodal observation $o_t = (v_t, s_t)$, and an interaction history $h_t^K$ that compensates for partial observability. It then samples an action from the policy
\begin{equation}
    a_t \sim \pi\left(\cdot \mid D, o_t, h_t^K\right),
\end{equation}
where $h_t^K = \{(o_i, a_i)\}_{i=t-K}^{t-1}$ comprises the $K$ most recent observation--action pairs. 
Based on the ablation study in Sec.~\ref{sec:exp-ablation}, we set the history window to $K^* = 1$. The environment then executes the sampled action $a_t$ until the next pause, transitioning the game state to $z_{t+1} = \mathcal{T}(z_t, a_t)$ and returning the next observation $o_{t+1}$. This interaction continues until the game terminates or the step budget is exhausted at step $H$, forming a trajectory $\tau = \{(o_t, a_t)\}_{t=1}^{H}$, to which a task evaluator assigns a binary outcome $R(\tau) \in \{0, 1\}$.}

\rev{\textbf{Action space.}
Each action in $\mathcal{A}$ is a native keyboard or mouse function call. Keyboard and mouse-button actions hold, tap, or release a key or button. Cursor actions move or drag to a target position given in normalized screenshot coordinates. 
A held key or button stays pressed across subsequent steps until it is released.
The agent can therefore emit a no-op to keep all inputs unchanged, which we refer to as a \emph{waiting step}. This action space is shared across all games, with its full schema given in Appendix~\ref{sec:unified-action}.}

\rev{\paragraph{Game environments.}
We select nine browser games spanning four genres. The detailed gameplay descriptions are provided in Appendix~\ref
{sec:game-descriptions}.
Each game defines \todo{five} tasks of increasing difficulty (\todo{45} tasks in total), with every episode initialized from a distinct random seed.
Each task includes an automated evaluator that checks whether an episode succeeds.
Each environment also exposes its internal game state (e.g., entity positions and scores) via a game API. The VLM agent, however, has no access to this state during gameplay and must infer it from multimodal observations.
}
% !TeX root = ../main.tex

\begin{figure*}[t]
    \centering
    \includegraphics[width=0.95\textwidth]{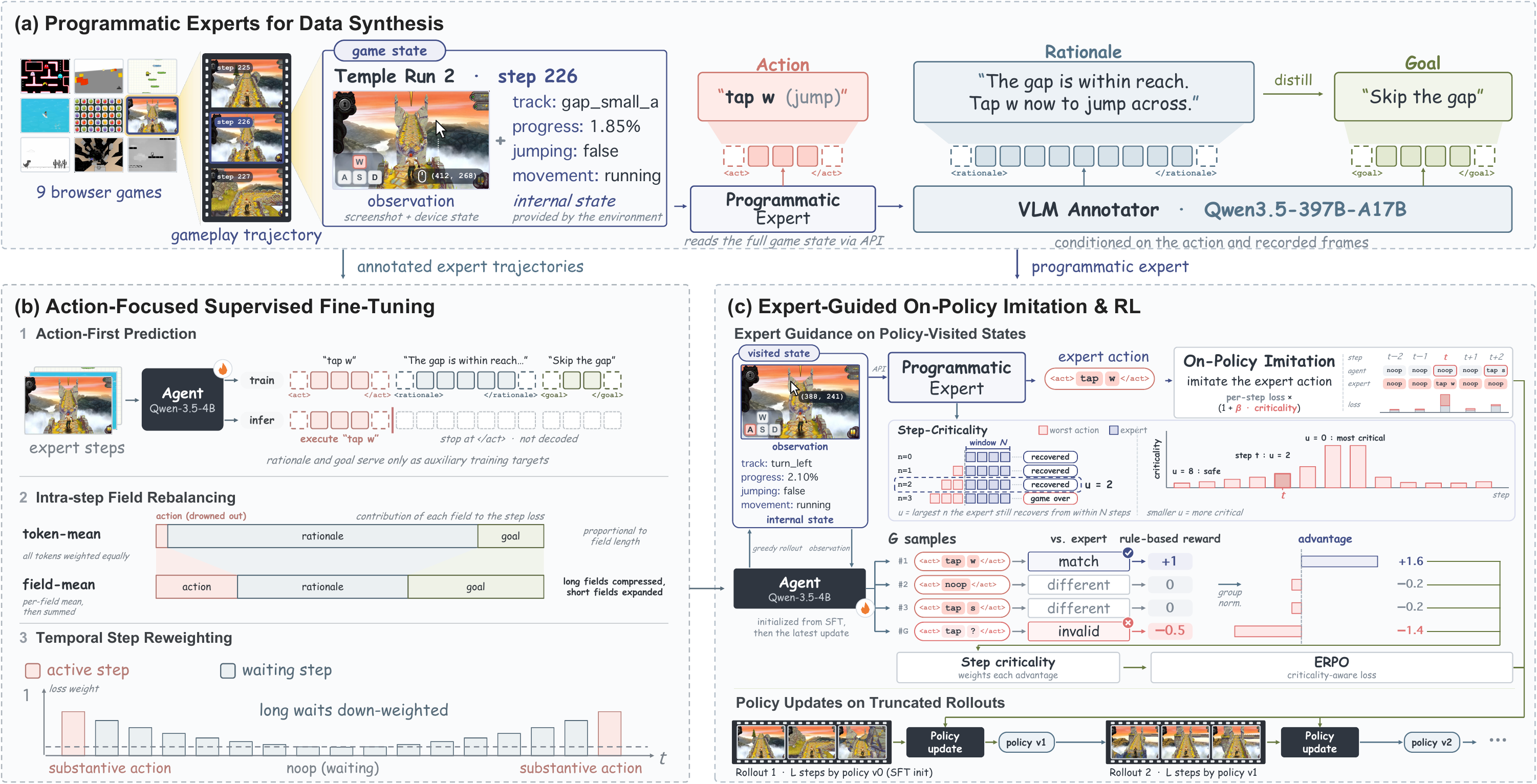}
    \caption{\justifying Overview of \tool. (a) Programmatic experts read game states to synthesize trajectories, which a VLM annotates with~rationales and goals. (b) Action-focused SFT cold-starts the agent with action-first prediction, intra-step field rebalancing, and temporal~step reweighting. (c) Over truncated rollouts, the expert supervises the agent on visited states via criticality-weighted on-policy imitation and~ERPO.}
    \label{fig:overview}
    \vspace{-1em}
\end{figure*}

\section{\tool}\label{sec:approach}\label{others}

We propose \tool, a three-stage framework (Fig.~\ref{fig:overview}) for training a game agent that interprets multimodal observations and acts through keyboard and mouse to complete tasks across multiple games.
First, we build a programmatic expert for each game and use it to generate trajectories for training (Sec.~\ref{sec:experts}).
Second, we cold-start the agent with supervised fine-tuning (SFT) on these trajectories, through which it acquires basic gameplay skills by imitating the expert (Sec.~\ref{sec:sft}). Third, to enable generalization beyond states covered by expert trajectories, we introduce expert-guided on-policy imitation and reinforcement learning, in which the programmatic expert supervises the agent on states visited in the agent's own rollouts (Sec.~\ref{sec:rl}).

\begin{table}[b]
\centering
\begin{revblock}
\caption{\rev{\justifying VLM agent (Qwen3.5-397B-A17B) vs.\ programmatic experts for training data collection. 
For each task, rollouts terminate once 25 successful trajectories are collected or 50 episodes have been attempted. 
\textit{TC} counts the tasks, out of five, with at least one successful trajectory.
\textit{SR} is the overall ratio of successful trajectories to attempted episodes across all tasks.
\textit{Tokens} is the average number of tokens consumed per successful trajectory.}}
\label{tab:data-source}
\footnotesize
\setlength{\tabcolsep}{4.8pt}
\begin{tabular}{@{}lcccccc@{}}
\toprule
 & \multicolumn{3}{c}{\textbf{VLM agent}} & \multicolumn{3}{c}{\textbf{Programmatic expert}} \\
\cmidrule(lr){2-4}\cmidrule(l){5-7}
{\scriptsize\textbf{Game}} & {\scriptsize\textbf{TC}} & {\scriptsize\textbf{SR}} & {\scriptsize\textbf{Tokens}} & {\scriptsize\textbf{TC}} & {\scriptsize\textbf{SR}} & {\scriptsize\textbf{Tokens}} \\
% \textbf{Game} & \textbf{Task Cov.} & \textbf{Succ.} & \textbf{Tokens} & \textbf{Task Cov.} & \textbf{Succ.} & \textbf{Tokens} \\
\midrule
Pac-Man         & \todo{2 / 5} & \todo{7.6} & \todo{53.7M} & \todo{5 / 5} & \todo{100.0} & \todo{2.8M} \\
Doodle Jump     & \todo{1 / 5} & \todo{10.7} & \todo{4.7M} & \todo{5 / 5} & \todo{100.0} & \todo{4.2M} \\
Orchard Match-3 & \todo{5 / 5} & \todo{80.1} & \todo{2.2M} & \todo{5 / 5} & \todo{96.2} & \todo{0.1M} \\
Chrome Dino     & \todo{0 / 5} & \todo{0.0} & \todo{--} & \todo{5 / 5} & \todo{74.4} & \todo{7.4M} \\
Cubefield       & \todo{0 / 5} & \todo{0.0} & \todo{--} & \todo{5 / 5} & \todo{100.0} & \todo{7.6M} \\
Edge Surf       & \todo{4 / 5} & \todo{21.9} & \todo{8.4M} & \todo{5 / 5} & \todo{76.7} & \todo{6.6M} \\
NS-Shaft        & \todo{3 / 5} & \todo{14.6} & \todo{4.4M} & \todo{5 / 5} & \todo{100.0} & \todo{3.0M} \\
Run 3           & \todo{1 / 5} & \todo{9.3} & \todo{5.5M} & \todo{5 / 5} & \todo{89.9} & \todo{6.2M} \\
Temple Run 2    & \todo{0 / 5} & \todo{0.0} & \todo{--} & \todo{5 / 5} & \todo{100.0} & \todo{9.3M} \\
\midrule
\textbf{Overall} & \textbf{\todo{16/45}} & \todo{16.0} & \todo{13.2M} & \textbf{\todo{45/45}} & \todo{93.0} & \todo{5.2M} \\
\bottomrule
\end{tabular}
\end{revblock}
\end{table}

\subsection{Programmatic Experts for Data Synthesis}\label{sec:experts}

To equip the agent with the fundamental game-playing capabilities required by the tasks in Sec.~\ref{sec:preliminaries}, we first collect successful trajectories as training supervision. Toward efficient data collection, we investigate two pipelines: a VLM agent and programmatic experts.

\textbf{VLM agent pipeline.}
We first construct a VLM-based agent to collect these trajectories. At each decision step $t$, the agent is conditioned on the context $D$, the current observation $o_t$, and the interaction history $h_t^K$, and predicts the next action along with the corresponding rationale and goal. The complete agent prompt is provided in Appendix~\ref{sec:agent-prompts}.
To decouple inference latency from the game dynamics, the environment is paused during inference and then advanced by $50\,\mathrm{ms}$ under the predicted action before the next decision step, allowing the agent to respond promptly to fast-moving objects.
Each rollout of a task uses a distinct game seed to ensure trajectory diversity, and we retain only the trajectories verified as successful by the task evaluator.
Nevertheless, even with Qwen3.5-397B-A17B as the underlying VLM, the rollout success rate is only 16.0\%, while failed rollouts still incur token consumption (see Table~\ref{tab:data-source}).

\textbf{Programmatic expert pipeline.}
We therefore replace the VLM in the same loop with a programmatic expert designed separately for each game.
Each expert is a parameter-free heuristic policy derived from the rules and logic of its game; the expert definitions for all games are given in Appendix~\ref{sec:expert-models}.
At each step $t$, the expert reads the internal game state through the game API and selects the best keyboard and mouse action for the current state. 
The resulting trajectories pair each observation $o_t$ with the expert action $a_t$, yet lack step-level rationales and goals.
To fill this gap, we prompt Qwen3.5-397B-A17B to annotate each step offline with a rationale and a goal, conditioned on the preceding steps, the expert action $a_t$, and the screenshots $v_t$ and $v_{t+1}$ before and after it. The annotation prompt is given in Appendix~\ref{sec:annotation-prompt}.

As shown in Table~\ref{tab:data-source}, the programmatic experts raise the rollout success rate over the VLM-based agent by \todo{77.0\%} and reduce the average token cost per successful trajectory by \todo{2.5}$\times$. 
Whereas the VLM agent consumes tokens on failed rollouts, the programmatic experts discard failures at zero token cost and incur tokens only for annotating successful trajectories.
We therefore adopt the programmatic experts to collect \todo{25} successful trajectories per task, yielding a dataset $\mathcal{D}_E$ of \todo{598.7K} annotated steps across \todo{1,125} trajectories.

% \begin{table}[b]
% \centering
% \small
% \caption{\justifying \textcolor{blue}{xxx}Success rate (\%) of xx,  (Level 1 is the easiest). Each overall entry aggregates 360 rollouts and each level 72 rollouts.}
% \label{tab:output-format}
% \begin{tabular}{lcccccc}
% \toprule
% Format & Overall & L1 & L2 & L3 & L4 & L5 \\
% \midrule
% \textsc{act-only}     & -- & -- & -- & -- & -- & -- \\
% \textsc{reason-first} & -- & -- & -- & -- & -- & -- \\
% \textsc{act-first}    & -- & -- & -- & -- & -- & -- \\
% \bottomrule
% \end{tabular}
% \end{table}

\subsection{Action-Focused Supervised Fine-Tuning}\label{sec:sft}
\looseness=-1 Given the annotated expert trajectories, we cold-start the policy $\pi$ via \emph{action-focused SFT}. To leverage auxiliary annotations without letting verbose rationales and repetitive waiting steps dilute action supervision, we introduce three designs.

\textbf{Action-First Prediction.}
\looseness=-1 Since only the action is executed during gameplay, we first determine how to structure the action, rationale, and goal in the policy output to best support action prediction.
To this end, we compare three output formats: \textsc{act-only} ($a$), \textsc{reason-first} ($r \oplus g \oplus a$), and \textsc{act-first} ($a \oplus r \oplus g$), where $\oplus$ denotes concatenation via special tokens, $r$ is the detailed rationale for executing $a$, and $g$ is the key goal distilled from $r$.
As shown in Table~\ref{tab:ablation-output-format}, \textsc{act-first} achieves the highest average success rate across all games.

\noindent\textbf{Observation 1.} \textit{Auxiliary supervision on the rationale and goal benefits action learning on average, especially when the action is predicted first.}

\looseness=-1 We therefore adopt \textsc{act-first}, serializing each target as \texttt{<act>}, \texttt{<rationale>}, \texttt{<goal>}, and stop decoding at \texttt{</act>} during inference, so that the rationale and goal serve only as auxiliary targets. 
With input $x_t = (D, o_t, h_t^K)$ and target $y_t = a_t \oplus r_t \oplus g_t$, the policy is reformulated as
\begin{equation}
    \pi\bigl(y_{t,\le J} \mid x_t\bigr)
    = \prod_{j=1}^{J} \pi\bigl(y_{t,j} \mid x_t, y_{t,<j}\bigr),
    \label{eq:agent-generation}
\end{equation}
\looseness=-1 where $J = |y_t|$ in training and $J$ marks \texttt{</act>} otherwise.

% \begin{figure}[t]
% \centering
% \includegraphics[width=\linewidth]{fig/field_normalized_loss.pdf}
% \caption{\justifying Token-mean vs. field-mean loss. The token-mean loss weights every token equally, which leaves the action with only 1.4\% of the supervision. The field-mean loss averages the loss within each field and then sums over fields, so the action, rationale, and goal receive equal weight.}
% \label{fig:field-loss}
% \end{figure}

\captionsetup[subfigure]{font=small, skip=-5pt, justification=centering}
\captionsetup{justification=centering, singlelinecheck=false, skip=3pt}
\begin{figure}[t]
    \centering
    \includegraphics[width=\linewidth]{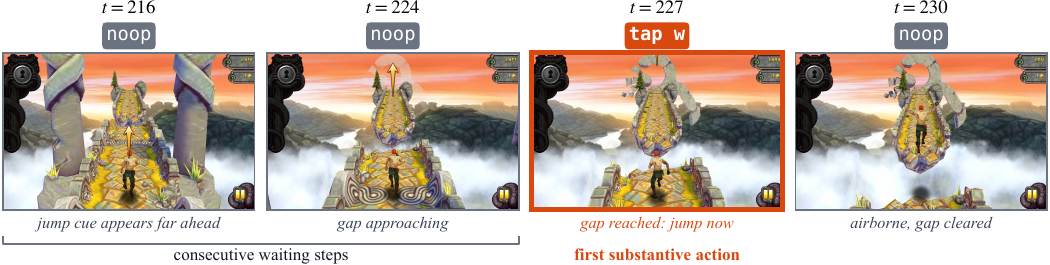}
    \caption{\justifying Long waiting sequence in Temple Run~2. The agent outputs no-ops while approaching a trap and takes an active step only at the critical frame just before it, pressing a key to jump.}
    \label{fig:long-wait}
    \vspace{-1em}
\end{figure}

\textbf{Intra-step Field Rebalancing.}
Within each step, the action is far shorter than the rationale and goal, accounting for only 1.4\% of target tokens across the training set.
Under the standard token-mean loss, each field contributes in proportion to its length, severely under-weighting the action.
We therefore propose an intra-step field-rebalancing (IFR) loss that weights all fields equally regardless of length.
Specifically, let $\mathcal F=\{\text{act},\text{rationale},\text{goal}\}$ denote the set of output fields, and let $I_{t,f}$ be the token positions occupied by field $f\in\mathcal F$ in the target $y_t$. The IFR loss at step $t$ is defined as
\begin{equation}
\ell_t^{\mathrm{IFR}} = -\sum_{f\in\mathcal F}\frac{1}{|I_{t,f}|}\sum_{j\in I_{t,f}}
    \log \pi(y_{t,j}\mid x_t,y_{t,<j}),
\label{eq:field-loss}
\end{equation}
which serves as the per-step loss for SFT.

% \begin{figure}[t]
% \centering
% \includegraphics[width=\linewidth]{fig/temporal_action_reweighting.pdf}
% \caption{\justifying Temporal step reweighting. Action steps keep unit weight, whereas the weight of a waiting step decays with its distance to the nearer action step until it reaches a floor, which down-weights the middle of a long wait.}
% \label{fig:temporal-weight}
% \end{figure}

\textbf{Temporal Step Reweighting.}
Decision steps fall into two types: \emph{active steps}, which change the device state through a substantive action (e.g., a key press), and \emph{waiting steps}, which leave it unchanged through a no-op (Fig.~\ref{fig:long-wait}). Waiting steps account for 93.2\% of the collected data, so weighting all steps equally lets them dominate the loss, causing an inter-step imbalance that biases the model toward waiting even when an active step is required. 
To mitigate this imbalance, we compare two strategies (Fig.~\ref{fig:temporal_decay}, left): \textsc{random-drop}, which randomly drops waiting steps to a 1:1 ratio with active steps, and \textsc{temporal-decay}, which retains all steps while decaying waiting-step weights with distance from the nearest active step. As shown in Fig.~\ref{fig:temporal_decay} (right), \textsc{temporal-decay} consistently outperforms \textsc{random-drop}.

\newsavebox{\tempdecayleftbox}
\newsavebox{\tempdecayrightbox}
\begin{figure}[t]
  \centering
  \sbox{\tempdecayleftbox}{\includegraphics[width=0.63\linewidth]{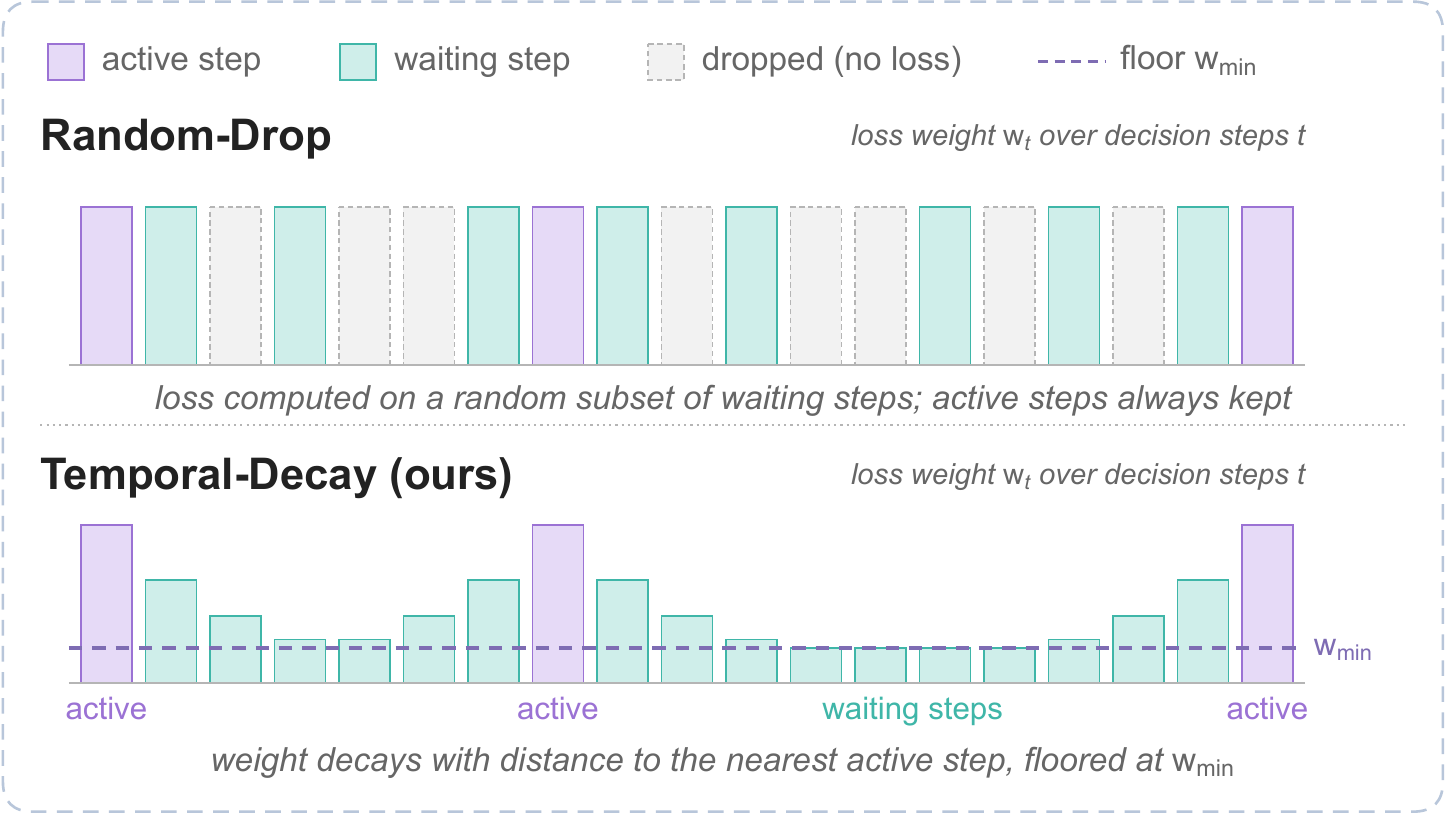}}
  \sbox{\tempdecayrightbox}{\includegraphics[width=0.34\linewidth]{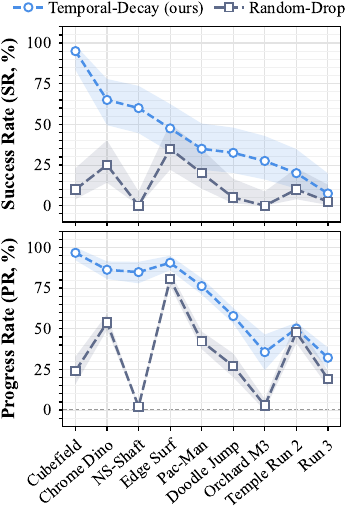}}
  \begin{minipage}[b]{0.63\linewidth}
      \centering
      \raisebox{\dimexpr(\ht\tempdecayrightbox-\ht\tempdecayleftbox)/2\relax}[\ht\tempdecayrightbox][0pt]{\usebox{\tempdecayleftbox}}
  \end{minipage}%
  \hfill
  \begin{minipage}[b]{0.34\linewidth}
      \centering
      \usebox{\tempdecayrightbox}
  \end{minipage}
  \caption{\justifying Comparison of step-weighting strategies. Left: \textsc{random-drop} drops waiting steps to a 1:1 ratio with active steps, whereas \textsc{temporal-decay} keeps all steps and decays waiting-step weights with distance from the nearest active step. Right: across all nine games, \textsc{temporal-decay} outperforms \textsc{random-drop} by 31.4\% in SR and 34.5\% in PR on average.}
  \label{fig:temporal_decay}
  \vspace{-1em}
\end{figure}

\noindent\textbf{Observation 2.} \textit{Waiting steps near active steps are crucial for action timing, whereas distant ones are less informative.}

\looseness=-1 We thus adopt \textsc{temporal-decay}. Specifically, we partition each trajectory into action segments, each starting at an active step and spanning the waiting steps until the next one.
Each step $t$ thus falls into exactly one segment; let $m_t$ be the length of that segment and $i_t\in\{0,\ldots,m_t-1\}$ the position of $t$ within it. We assign the temporal weight
\begin{equation}
w_t = \max\!\left\{w_{\min},\,\gamma^{\,i_t},\,\gamma^{\,m_t-i_t}\right\},
\label{eq:run-weight}
\end{equation}
\looseness=-1 where $\gamma$ is the decay rate and $w_{\min}$ is the weight floor, leaving unit weight to active steps ($i_t=0$). Weighting $\ell_t^{\mathrm{IFR}}$ by $w_t$ normalized by its dataset mean $\bar w$ to preserve the loss scale, we define the action-focused SFT objective over $\mathcal{D}_E$ as
\begin{equation}
\mathcal L_{\mathrm{AFT}} =
  \frac{1}{|\mathcal{D}_E|}\sum_{t\in\mathcal{D}_E}
  \frac{w_t}{\bar w}\,\ell_t^{\mathrm{IFR}}.
\label{eq:sft-loss}
\end{equation}

\subsection{Expert-Guided On-Policy Learning}\label{sec:rl}

\looseness=-1 Supervising the policy solely on offline expert trajectories leaves the agent prone to compounding errors upon encountering unseen game states.
On-policy training on the agent's own rollouts mitigates this distribution shift, yet sparse terminal rewards hinder effective credit assignment over hundreds of steps.
To address this, we introduce an expert-guided process reward that densifies the learning signal at the step level. This dense reward allows us to partition each trajectory into segments and update the policy segment by segment progressively. Specifically, expert guidance operates in three ways: (i) a step-criticality measure quantifies how consequential each step is (Sec.~\ref{sec:step-urgency}); (ii) on-policy imitation aligns the agent's visited states with expert actions (Sec.~\ref{sec:on-policy-imitation}); and (iii) expert-rewarded policy optimization updates the policy under step-level expert rewards (Sec.~\ref{sec:expert-guided-grpo}). The step-criticality from (i) reweights the per-step objectives in (ii) and (iii), concentrating learning on the most decision-critical steps.

\subsubsection{Decision-Step Criticality Measure}\label{sec:step-urgency}

\looseness=-1 Not all steps along a trajectory are equally important: at some states, deviating from the expert action $a^{E}_t$ is recoverable, as the environment is somewhat fault-tolerant, whereas at others a single misstep leads to irreversible failure. We quantify this difference in sensitivity as the step criticality. Concretely, at step $t$, we first define the worst action $\bar{a}_t$ as the rule-based inversion of the expert action $a^{E}_t$ (e.g., if $a^{E}_t$ presses a key, then $\bar{a}_t$ refrains from pressing it). 
Given a horizon $N$, the slack $\sigma_t\in\{0,\ldots,N\}$ is defined as the largest $n\le N$ satisfying the following condition: 
after simulating $\bar{a}_t$ for $n$ consecutive steps and then letting the expert take over for a fixed time window, the episode survives without termination (i.e., the deviation remains recoverable, and unrecoverable otherwise).
The slack $\sigma_t$ thus measures the maximum number of consecutive mistakes a step can tolerate: the smaller $\sigma_t$, the less room for error, and the more critical the step. Accordingly, we define the criticality weight $\kappa_t$ as:
\begin{equation}
\kappa_t=1+\beta c_t,\quad
c_t=\frac{1}{1+\sigma_t},
\label{eq:opi-weight}
\end{equation}
where $\beta$ controls the weighting strength, and $c_t=0$ if the state is already unrecoverable at $n=0$.

\subsubsection{On-Policy Imitation}\label{sec:on-policy-imitation}
On-policy imitation trains the agent to predict the expert's action on the states that the agent itself visits. 
In round $k$ (with $\pi_0$ initialized as the SFT policy), these states are produced by greedily rolling out the policy $\pi_k$, and the expert action $a_t^E$ serves as the supervision target at each step $t$. 
Since steps differ in criticality, we adopt criticality-adaptive imitation, which strictly supervises critical steps while relaxing non-critical ones. 
This relaxation also mitigates entropy collapse to some extent, since strictly imitating a deterministic expert at every step would make the policy increasingly homogeneous. Moreover, because the expert labels $a_t^E$ remain dominated by waiting steps, we also compute the temporal weight $w_t$ (Eq.~\eqref{eq:run-weight}) over the $a_t^E$ sequence.
Combining both into the step weight $b_t=w_t\,\kappa_t$, the imitation loss at step $t$ is
\begin{equation}
\ell_t^{\mathrm{OPI}}=
  -\frac{b_t}{\bar b}\sum_{j=1}^{|a_t^E|}
  \log \pi\bigl(a_{t,j}^{E}\mid x_t,a_{t,<j}^{E}\bigr),
\label{eq:opi-loss}
\end{equation}
where $a_{t,j}^{E}$ denotes the $j$-th token of $a_t^E$, and $\bar b$ is the mean of $b_t$ across steps in each update, preserving the loss scale.

\subsubsection{Expert-Rewarded Policy Optimization}\label{sec:expert-guided-grpo}
Alongside the imitation loss, we further introduce Expert-Rewarded Policy
Optimization (ERPO), which rewards each action by whether it matches the expert
action $a_t^E$. At each step $t$, the policy $\pi_k$ samples a group of $G$ candidate actions
$a_t^{(i)}\sim\pi_k(\cdot\mid x_t)$, and the reward of candidate $i$ is
\begin{equation}
r_t^{(i)} =
\begin{cases}
1 & \text{if } a_t^{(i)}\simeq a_t^E,\\
-0.5 & \text{if } a_t^{(i)} \text{ is invalid},\\
0 & \text{otherwise},
\end{cases}
\label{eq:expert-reward}
\end{equation}
where $\simeq$ means that two actions have the same effect on the input devices, and an action is invalid if it is malformed or uses controls that the game does not allow.

Standardizing rewards within each group places every step on the same scale regardless of its criticality. We therefore introduce a criticality-adaptive advantage that scales candidate $i$'s standardized reward by $\kappa_t/\bar\kappa$ (Eq.~\eqref{eq:opi-weight}):
\begin{equation}
A_t^{(i)}=\frac{\kappa_t}{\bar\kappa}\cdot
  \frac{r_t^{(i)}-\operatorname{mean}(\mathbf r_t)}{\operatorname{std}(\mathbf r_t)+\delta},
\label{eq:criticality-advantage}
\end{equation}
where $\mathbf r_t=(r_t^{(1)},\ldots,r_t^{(G)})$, $\delta$ is a small constant, and $\bar\kappa$ is the mean of $\kappa_t$.
Then, the ERPO loss at step $t$ is
\begin{align}
&\ell_t^{\mathrm{ERPO}}=
  -\sum_{i=1}^{G}\sum_{j=1}^{|a_t^{(i)}|}
  \min\Biggl(
  \frac{\pi\bigl(a_{t,j}^{(i)}\mid x_t,a_{t,<j}^{(i)}\bigr)}
       {\pi_k\bigl(a_{t,j}^{(i)}\mid x_t,a_{t,<j}^{(i)}\bigr)}A_t^{(i)},
  \notag\\
&\quad\operatorname{clip}\Biggl(
  \frac{\pi\bigl(a_{t,j}^{(i)}\mid x_t,a_{t,<j}^{(i)}\bigr)}
       {\pi_k\bigl(a_{t,j}^{(i)}\mid x_t,a_{t,<j}^{(i)}\bigr)},
  \,1-\epsilon,\,1+\epsilon\Biggr)A_t^{(i)}\Biggr).
\label{eq:grpo-loss}
\end{align}

\subsubsection{Policy Updates on Truncated Rollouts}\label{sec:truncated-rollouts}
We update the policy on each truncated rollout of $L$ consecutive steps with the
criticality-adaptive losses, which supply per-step supervision. Specifically,
over the step set $\mathcal{B}_k$ collected by $\pi_k$ in one rollout,
we minimize the joint objective
\begin{equation}
\mathcal L_k=\frac{1}{Z_k}\sum_{t\in\mathcal B_k}
  \Bigl(\ell_t^{\mathrm{OPI}}
  +\ell_t^{\mathrm{ERPO}}\Bigr),
\label{eq:rl-loss}
\end{equation}
where $Z_k$ is the total number of action tokens in $\mathcal B_k$.
The updated policy $\pi_{k+1}$ then resumes the ongoing episode from the last visited state to collect $\mathcal B_{k+1}$, repeating this rollout-and-update cycle until the episode terminates.

% !TeX root = ../main.tex

\section{Experiments}\label{sec:experiments}

\subsection{Experimental Setup}\label{sec:exp-setup}

\textbf{Evaluation Environment.}
We evaluate visual game agents in online gameplay across the nine games and \todo{45} tasks introduced in Sec.~\ref{sec:preliminaries}.
For each task, the agent attempts 8 episodes with distinct game seeds that are strictly disjoint from those used in expert trajectory collection and training.
All episodes run in a headless Chromium browser with a viewport of $1280\times800$ pixels. The detailed training, decoding, and hardware settings are provided in Appendix~\ref{sec:implementation-details}.

\textbf{Baselines.}
We compare \tool against four large vision-language models: Qwen3.5-397B-A17B, DeepSeek-V4.1-Flash, GLM-5.3, and Kimi-K2.6.
All baselines and our trained model share the same agent scaffold, with an identical system prompt, input device state, history window, and action tools (Appendix~\ref{sec:agent-prompts}), and follow the same evaluation protocol for a fair and reproducible comparison.

\textbf{Evaluation metrics.}
For each game, we compute three metrics over all episodes:
\textbf{Success Rate (SR)}, the fraction of episodes in which the agent fully achieves the task goal within the step budget;
\textbf{Progress Rate (PR)}, the average task progress across episodes, where each episode's progress is the fraction of the task goal it reaches;
and \textbf{Task Coverage (TC)}, the fraction of tasks the agent solves at least once, i.e., a task is covered if any of its episodes succeeds.

% case stydy, 分位线的散点图，
\subsection{Main Results}\label{sec:exp-main}

\begin{table*}[t]
\centering
\caption{Main Results on the Nine Games.}
\label{tab:main-results}
\scriptsize
\setlength{\tabcolsep}{1.8pt}
\renewcommand{\arraystretch}{1.15}
\resizebox{\textwidth}{!}{%
\begin{tabular}{@{}l@{\hspace{6pt}}*{9}{C{1.85em}C{1.85em}C{1.85em}@{\hspace{7pt}}}*{3}{>{\columncolor{gray!10}}C{1.85em}}@{}}
\toprule
\multirow{2}{*}{\textbf{Model}}
& \multicolumn{3}{c}{\textbf{Pac-Man}}
& \multicolumn{3}{c}{\textbf{Doodle Jump}}
& \multicolumn{3}{c}{\textbf{Orchard Match-3}}
& \multicolumn{3}{c}{\textbf{Chrome Dino}}
& \multicolumn{3}{c}{\textbf{Cubefield}}
& \multicolumn{3}{c}{\textbf{Edge Surf}}
& \multicolumn{3}{c}{\textbf{NS-Shaft}}
& \multicolumn{3}{c}{\textbf{Run 3}}
& \multicolumn{3}{c}{\textbf{Temple Run 2}}
& \multicolumn{3}{c}{\cellcolor{gray!10}\textbf{Overall}} \\
\cmidrule(lr){2-4} \cmidrule(lr){5-7} \cmidrule(lr){8-10} \cmidrule(lr){11-13} \cmidrule(lr){14-16}
\cmidrule(lr){17-19} \cmidrule(lr){20-22} \cmidrule(lr){23-25} \cmidrule(lr){26-28} \cmidrule(lr){29-31}
& SR & PR & TC & SR & PR & TC & SR & PR & TC
& SR & PR & TC & SR & PR & TC & SR & PR & TC
& SR & PR & TC & SR & PR & TC & SR & PR & TC
& SR & PR & TC \\
\midrule
Qwen3.5-397B-A17B
  & 7.5 & 46.5 & 40.0 & 15.0 & 38.5 & 20.0 & 92.5 & 95.0 & 100.0 & 0.0 & 26.3 & 0.0 & 0.0 & 15.7 & 0.0
  & 22.5 & 65.6 & 80.0 & 15.0 & 57.2 & 60.0 & 10.0 & 31.1 & 20.0 & 0.0 & 8.3 & 0.0 & 18.1 & 42.7 & 35.6 \\
DeepSeek-V4.1-Flash
  & 20.0 & 58.0 & 40.0 & 17.5 & 35.7 & 40.0 & 0.0 & 0.6 & 0.0 & 0.0 & 22.2 & 0.0 & 5.0 & 48.3 & 20.0
  & 35.0 & 74.0 & 80.0 & 10.0 & 50.8 & 20.0 & 0.0 & 21.7 & 0.0 & 5.0 & 30.0 & 40.0 & 10.3 & 37.9 & 26.7 \\
GLM-5.3
  & 0.0 & 2.0 & 0.0 & 72.5 & 78.0 & 80.0 & 0.0 & 0.0 & 0.0 & 0.0 & 15.8 & 0.0 & 0.0 & 4.2 & 0.0
  & 0.0 & 0.4 & 0.0 & 7.5 & 23.4 & 20.0 & 0.0 & 35.1 & 0.0 & 0.0 & 13.8 & 0.0 & 8.9 & 19.2 & 11.1 \\
Kimi-K2.6
  & 12.5 & 19.7 & 40.0 & 0.0 & 0.0 & 0.0 & 0.0 & 0.0 & 0.0 & 0.0 & 21.7 & 0.0 & 0.0 & 10.3 & 0.0
  & 20.0 & 26.8 & 40.0 & 27.5 & 30.4 & 40.0 & 12.5 & 28.4 & 20.0 & 27.5 & 57.3 & 40.0 & 11.1 & 21.6 & 20.0 \\
\midrule
\rowcolor{blue!6}
\textbf{\tool (ours)}
  & 77.5 & 98.6 & 100.0 & 45.0 & 82.8 & 80.0 & 45.0 & 48.8 & 100.0 & 90.0 & 95.1 & 100.0 & 82.5 & 89.9 & 100.0
  & 60.0 & 93.9 & 60.0 & 90.0 & 93.7 & 100.0 & 12.5 & 36.2 & 20.0 & 17.5 & 93.0 & 20.0 & 57.8 & 81.3 & 75.6 \\
\bottomrule
\end{tabular}%
}
\end{table*}

\textbf{Overall performance.} Table~\ref{tab:main-results} reports the success rate, progress rate, and task coverage of all agents across the nine games.
\tool achieves 57.8\% SR, 81.3\% PR, and 75.6\% TC.
It outperforms the strongest baseline, Qwen3.5-397B-A17B (18.1\% SR, 42.7\% PR, and 35.6\% TC), by 39.7\% in SR, 38.6\% in PR, and 40.0\% in TC.
Per-game results reveal three key insights.
First, general-purpose VLMs perform reasonably on isolated games, such as Qwen3.5-397B-A17B on static puzzle matching in Orchard Match-3 (92.5\% SR) and GLM-5.3 on Doodle Jump (72.5\% SR), yet they fail to generalize across diverse game mechanics.
Second, fast-paced games requiring precise action timing, such as Chrome Dino and Cubefield, remain major bottlenecks for general VLMs, where every baseline scores at most 5.0\% SR.
In contrast, \tool reaches 90.0\% SR on Chrome Dino and 82.5\% SR on Cubefield, achieving 100\% TC on five out of nine games.
Third, even on challenging 3D runners like Temple Run 2, \tool attains 93.0\% PR despite a 17.5\% SR, showing that it maintains stable long-horizon control.

\begin{figure}[!htb]
\centering
\includegraphics[width=\linewidth]{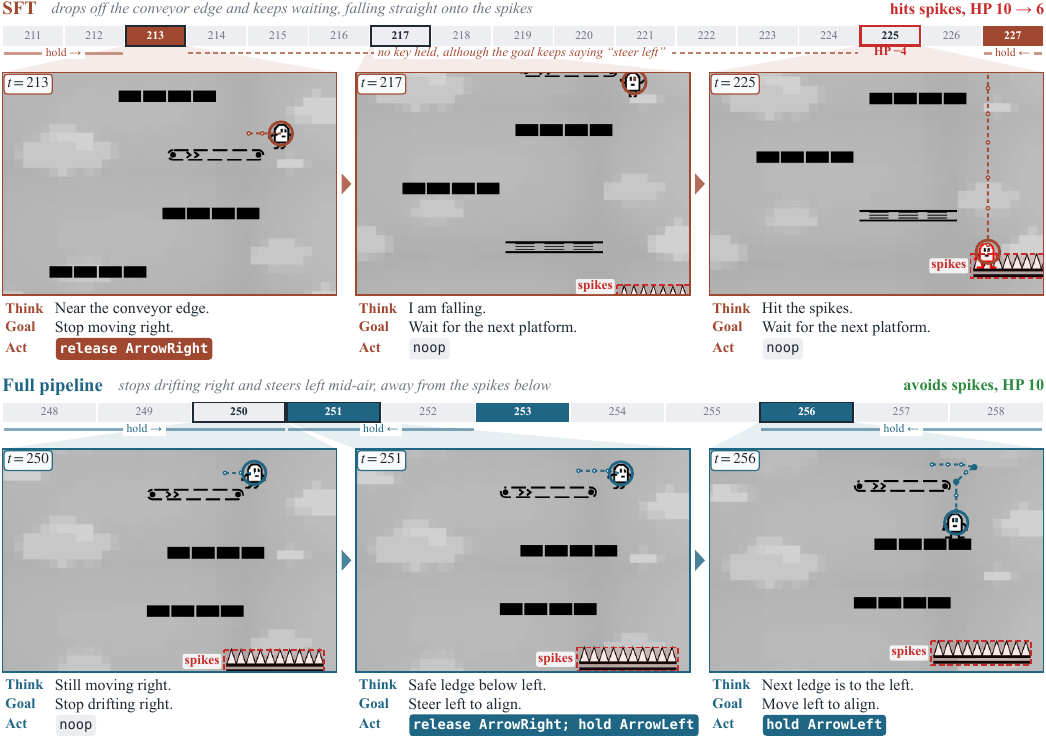}
\caption{\justifying Case study on NS-Shaft. Top: the SFT policy drops off the conveyor and keeps waiting, falling onto the spikes. Bottom: after expert-guided on-policy learning, the agent steers left in mid-air and lands on the safe ledge.}
\label{fig:case-study}
\vspace{-1em}
\end{figure}

\textbf{Case Study.}
Fig.~\ref{fig:case-study} compares the action-focused SFT policy and the full \tool pipeline on NS-Shaft, where a rightward conveyor carries the character toward an edge above spikes.
Although the SFT policy releases \texttt{ArrowRight} at the edge ($t{=}213$), it keeps outputting \texttt{noop} while falling and lands directly on the spikes ($t{=}225$).
In contrast, the agent trained by the full pipeline promptly releases \texttt{ArrowRight} and holds \texttt{ArrowLeft} in mid-air ($t{=}251$), landing safely on the left ledge ($t{=}256$).
Because such off-edge states rarely appear in offline expert trajectories, this contrast confirms that expert-guided on-policy learning teaches timely recovery in unseen states.

\subsection{Ablation Study}\label{sec:exp-ablation}

To evaluate the effectiveness of each component in \tool, we conduct comprehensive ablation studies. 

\begin{table}[!t]
\centering
\newcommand{\ablDown}[1]{{\tiny\textcolor{red!70!black}{$\downarrow$#1}}}
\newcommand{\ablUp}[1]{{\tiny\textcolor{green!45!black}{$\uparrow$#1}}}
\caption{Ablation on the Action-Focused SFT Output Format.}
\label{tab:ablation-output-format}
\footnotesize
\setlength{\tabcolsep}{2.2pt}
\renewcommand{\arraystretch}{1.05}
\resizebox{\columnwidth}{!}{%
\begin{tabular}{l*{3}{C{2.3em}}*{3}{C{3.5em}}*{3}{>{\columncolor{blue!6}}C{3.5em}}}
\toprule
& \multicolumn{3}{c}{\makecell{\textbf{\textsc{act-only}}\\{\color{gray}($a_t$)}}}
& \multicolumn{3}{c}{\makecell{\textbf{\textsc{reason-first}}\\{\color{gray}($r_t \oplus g_t \oplus a_t$)}}}
& \multicolumn{3}{>{\columncolor{blue!6}\centering\arraybackslash}p{\dimexpr10.5em+4\tabcolsep\relax}}{\makecell{\textbf{\textsc{act-first}}\\{\color{gray}($a_t \oplus r_t \oplus g_t$)}}} \\
\cmidrule(lr){2-4} \cmidrule(lr){5-7} \cmidrule(lr){8-10}
\textbf{Game} & SR & PR & TC & SR & PR & TC & SR & PR & TC \\
\midrule
Pac-Man         & 2.5 & 25.7 & 20.0 & 50.0 & 85.0 & 80.0 & 35.0 & 76.2 & 80.0 \\
Doodle Jump     & 0.0 & 25.9 & 0.0 & 32.5 & 59.8 & 40.0 & 32.5 & 57.8 & 40.0 \\
Orchard Match-3 & 30.0 & 39.6 & 40.0 & 10.0 & 11.3 & 20.0 & 27.5 & 35.6 & 60.0 \\
Chrome Dino     & 72.5 & 88.8 & 100.0 & 67.5 & 87.3 & 100.0 & 65.0 & 86.3 & 100.0 \\
Cubefield       & 52.5 & 77.8 & 100.0 & 37.5 & 55.4 & 100.0 & 95.0 & 96.7 & 100.0 \\
Edge Surf       & 25.0 & 81.1 & 40.0 & 42.5 & 84.3 & 80.0 & 47.5 & 90.6 & 100.0 \\
NS-Shaft        & 22.5 & 58.4 & 60.0 & 47.5 & 76.3 & 100.0 & 60.0 & 84.8 & 100.0 \\
Run 3           & 2.5 & 22.8 & 20.0 & 0.0 & 22.4 & 0.0 & 7.5 & 32.2 & 20.0 \\
Temple Run 2    & 0.0 & 29.4 & 0.0 & 10.0 & 53.3 & 20.0 & 20.0 & 50.1 & 20.0 \\
\midrule
\rowcolor{gray!10}
\textbf{Overall}
  & 23.1 & 49.9 & 42.2
  & 33.1\,\ablUp{10.0} & 59.5\,\ablUp{9.6} & 60.0\,\ablUp{17.8}
  & 43.3\,\ablUp{20.2} & 67.8\,\ablUp{17.9} & 68.9\,\ablUp{26.7} \\
\bottomrule
\end{tabular}%
}
\end{table}

\textbf{Action-focused SFT.}\label{sec:ablation-output}
We ablate the three key designs of action-focused SFT: action-first prediction, intra-step field rebalancing (IFR), and temporal step reweighting.
% All variants are trained on the same expert trajectories with identical hyperparameters, and the resulting SFT models are evaluated directly without subsequent online learning.
We first examine the output format by comparing \textsc{act-first} with the two alternatives introduced in Sec.~\ref{sec:sft}.
\textsc{act-only} discards the rationale and goal and supervises the action alone, testing whether auxiliary annotations benefit action prediction.
\textsc{reason-first} retains all three fields but generates the rationale and goal before the action, testing the effect of generation order.
Table~\ref{tab:ablation-output-format} reports the results.\looseness=-1

Overall, \textsc{act-only} yields the lowest performance, with 23.1\% SR, 49.9\% PR, and 42.2\% TC.
Incorporating rationale and goal supervision in \textsc{reason-first} raises SR by 10.0\% to 33.1\%, alongside gains of 9.6\% in PR and 17.8\% in TC.
Reversing the generation order in \textsc{act-first} further boosts performance to 43.3\% SR, 67.8\% PR, and 68.9\% TC, surpassing \textsc{act-only} by 20.2\%, 17.9\%, and 26.7\%, respectively.
These results show that rationale and goal annotations provide effective auxiliary supervision, whereas generating them first can mislead subsequent action prediction.
We therefore adopt \textsc{act-first} to retain this training benefit without extra inference latency.\looseness=-1

\begin{table}[!t]
\centering
\newcommand{\ablDown}[1]{{\tiny\textcolor{red!70!black}{$\downarrow$#1}}}
\newcommand{\ablUp}[1]{{\tiny\textcolor{green!45!black}{$\uparrow$#1}}}
\caption{Ablation on the Action-Focused SFT Loss.}
\label{tab:ablation-sft-loss}
\footnotesize
\setlength{\tabcolsep}{2.2pt}
\renewcommand{\arraystretch}{1.05}
\resizebox{\columnwidth}{!}{%
\begin{tabular}{l*{3}{>{\columncolor{blue!6}}C{2.3em}}*{9}{C{3.5em}}}
\toprule
& \multicolumn{3}{>{\columncolor{blue!6}\centering\arraybackslash}p{\dimexpr6.9em+4\tabcolsep\relax}}{\makecell{\textbf{Action-Focused}\\{\textbf{SFT}} {\color{gray}(ours)}}}
& \multicolumn{3}{c}{\makecell{\textbf{w/o IFR}\\{\color{gray}(token-mean)}}}
& \multicolumn{3}{c}{\makecell{\textbf{w/o Reweight.}\\{\color{gray}($w_t\equiv1$)}}}
& \multicolumn{3}{c}{\makecell{\textbf{w/o Floor}\\{\color{gray}($w_{\min}=0$)}}} \\
\cmidrule(lr){2-4} \cmidrule(lr){5-7} \cmidrule(lr){8-10} \cmidrule(lr){11-13}
\textbf{Game} & SR & PR & TC & SR & PR & TC & SR & PR & TC & SR & PR & TC \\
\midrule
Pac-Man         & 35.0 & 76.2 & 80.0 & 22.5 & 51.1 & 40.0 & 40.0 & 72.3 & 80.0 & 45.0 & 75.9 & 80.0 \\
Doodle Jump     & 32.5 & 57.8 & 40.0 & 10.0 & 36.4 & 20.0 & 27.5 & 55.2 & 60.0 & 37.5 & 61.9 & 60.0 \\
Orchard Match-3 & 27.5 & 35.6 & 60.0 & 22.5 & 35.7 & 40.0 & 5.0 & 8.3 & 40.0 & 2.5 & 2.5 & 20.0 \\
Chrome Dino     & 65.0 & 86.3 & 100.0 & 70.0 & 88.0 & 100.0 & 77.5 & 91.7 & 100.0 & 85.0 & 94.0 & 100.0 \\
Cubefield       & 95.0 & 96.7 & 100.0 & 47.5 & 71.9 & 100.0 & 30.0 & 58.2 & 60.0 & 52.5 & 66.2 & 100.0 \\
Edge Surf       & 47.5 & 90.6 & 100.0 & 30.0 & 77.5 & 60.0 & 50.0 & 87.5 & 100.0 & 42.5 & 88.5 & 80.0 \\
NS-Shaft        & 60.0 & 84.8 & 100.0 & 32.5 & 63.0 & 60.0 & 75.0 & 90.0 & 100.0 & 67.5 & 89.2 & 100.0 \\
Run 3           & 7.5 & 32.2 & 20.0 & 0.0 & 20.5 & 0.0 & 7.5 & 32.3 & 20.0 & 5.0 & 29.3 & 20.0 \\
Temple Run 2    & 20.0 & 50.1 & 20.0 & 2.5 & 34.0 & 20.0 & 17.5 & 47.5 & 40.0 & 2.5 & 27.5 & 20.0 \\
\midrule
\rowcolor{gray!10}
\textbf{Overall}
  & 43.3 & 67.8 & 68.9
  & 26.4\,\ablDown{16.9} & 53.1\,\ablDown{14.7} & 48.9\,\ablDown{20.0}
  & 36.7\,\ablDown{6.6} & 60.3\,\ablDown{7.5} & 66.7\,\ablDown{2.2}
  & 37.8\,\ablDown{5.5} & 59.4\,\ablDown{8.4} & 64.4\,\ablDown{4.5} \\
\bottomrule
\end{tabular}%
}
\end{table}

We then ablate the training loss design under the \textsc{act-first} format.
\textit{w/o IFR} replaces the intra-step field-rebalancing loss $\ell_t^{\mathrm{IFR}}$ in Eq.~\eqref{eq:field-loss} with the standard token-mean loss.
\textit{w/o Reweight.} disables temporal step reweighting by setting $w_t\equiv1$ in Eq.~\eqref{eq:sft-loss}.
\textit{w/o Floor} sets $w_{\min}=0$ in Eq.~\eqref{eq:run-weight}, assigning near-zero weight to distant waiting steps.
Table~\ref{tab:ablation-sft-loss} reports the results. 
Compared with Action-Focused SFT, removing IFR reduces overall SR, PR, and TC by 16.9\%, 14.7\%, and 20.0\%, respectively.
Removing temporal step reweighting lowers SR, PR, and TC by 6.6\%, 7.5\%, and 2.2\%, while removing the weight floor drops them by 5.5\%, 8.4\%, and 4.5\%.
These results confirm that IFR is crucial to prevent verbose rationales and goals from dominating concise action tokens.
Temporal step reweighting curbs bias toward passive waiting, and a nonzero weight floor preserves necessary supervision during long waits.

\begin{figure}[!t]
\centering
\includegraphics[width=\linewidth]{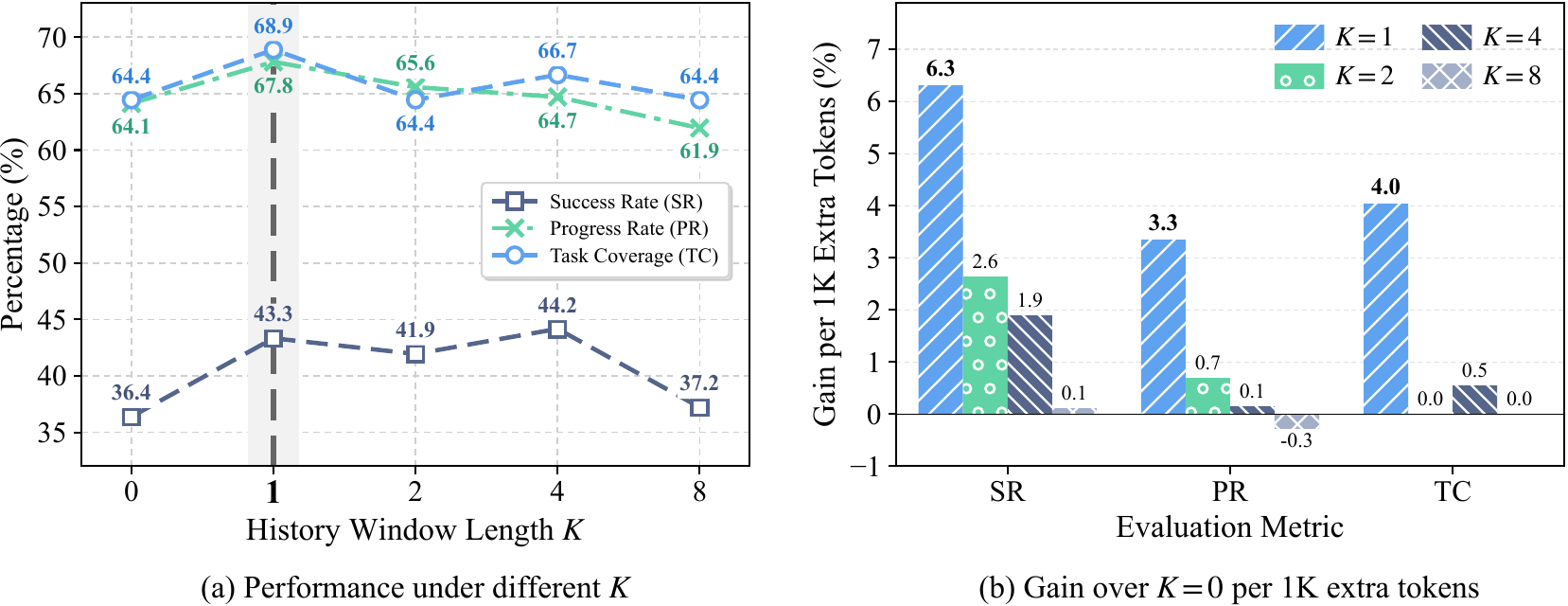}
\caption{\justifying Ablation on the History Window Length $K$.}
\label{fig:ablation-history-window}
\vspace{-0.5em}
\end{figure}

\textbf{History window.}
We ablate the history window length $K \in \{0, 1, 2, 4, 8\}$ during SFT (Fig.~\ref{fig:ablation-history-window}).
In Fig.~\ref{fig:ablation-history-window}(a), increasing $K$ from 0 to 1 boosts overall SR, PR, and TC from 36.4\%, 64.1\%, and 64.4\% to 43.3\%, 67.8\%, and 68.9\%, achieving the highest PR and TC across all settings.
Further increasing $K$ brings no consistent gains and markedly degrades performance at $K=8$.
Moreover, in Fig.~\ref{fig:ablation-history-window}(b), $K=1$ delivers the highest gain per 1K extra tokens across all metrics.
These results indicate that a single preceding step suffices to capture object motion, whereas longer histories add redundant context and token cost.
We therefore set $K=1$.\looseness=-1

\begin{table}[!t]
\centering
\newcommand{\ablDown}[1]{{\tiny\textcolor{red!70!black}{$\downarrow$#1}}}
\newcommand{\ablUp}[1]{{\tiny\textcolor{green!45!black}{$\uparrow$#1}}}
\caption{Ablation on Expert-Guided On-Policy Learning.}
\label{tab:ablation-online-model}
\footnotesize
\setlength{\tabcolsep}{2.2pt}
\renewcommand{\arraystretch}{1.05}
\resizebox{\columnwidth}{!}{%
\begin{tabular}{l*{3}{>{\columncolor{blue!6}}C{2.3em}}*{9}{C{3.5em}}}
\toprule
& \multicolumn{3}{>{\columncolor{blue!6}\centering\arraybackslash}p{\dimexpr6.9em+4\tabcolsep\relax}}{\makecell{\textbf{\tool}\\{\color{gray}(OPI + ERPO)}}}
& \multicolumn{3}{c}{\makecell{\textbf{w/o ERPO}\\{\color{gray}($\ell_t^{\mathrm{OPI}}$ only)}}}
& \multicolumn{3}{c}{\makecell{\textbf{w/o OPI}\\{\color{gray}($\ell_t^{\mathrm{ERPO}}$ only)}}}
& \multicolumn{3}{c}{\makecell{\textbf{w/o $\boldsymbol{\kappa_t}$}\\{\color{gray}($\kappa_t\equiv1$)}}} \\
\cmidrule(lr){2-4} \cmidrule(lr){5-7} \cmidrule(lr){8-10} \cmidrule(lr){11-13}
\textbf{Game} & SR & PR & TC & SR & PR & TC & SR & PR & TC & SR & PR & TC \\
\midrule
Pac-Man         & 77.5 & 98.6 & 100.0 & 37.5 & 83.4 & 40.0 & 22.5 & 58.9 & 40.0 & 40.0 & 75.4 & 60.0 \\
Doodle Jump     & 45.0 & 82.8 & 80.0 & 17.5 & 65.4 & 60.0 & 10.0 & 46.9 & 60.0 & 20.0 & 61.5 & 60.0 \\
Orchard Match-3 & 45.0 & 48.8 & 100.0 & 0.0 & 2.0 & 0.0 & 42.5 & 51.9 & 80.0 & 30.0 & 34.4 & 60.0 \\
Chrome Dino     & 90.0 & 95.1 & 100.0 & 0.0 & 19.4 & 0.0 & 62.5 & 77.4 & 100.0 & 57.5 & 71.6 & 100.0 \\
Cubefield       & 82.5 & 89.9 & 100.0 & 80.0 & 88.8 & 100.0 & 72.5 & 86.4 & 100.0 & 60.0 & 83.9 & 100.0 \\
Edge Surf       & 60.0 & 93.9 & 60.0 & 42.5 & 80.7 & 60.0 & 52.5 & 89.4 & 60.0 & 60.0 & 94.1 & 60.0 \\
NS-Shaft        & 90.0 & 93.7 & 100.0 & 65.0 & 85.7 & 100.0 & 82.5 & 94.8 & 100.0 & 70.0 & 86.6 & 100.0 \\
Run 3           & 12.5 & 36.2 & 20.0 & 5.0 & 28.0 & 20.0 & 10.0 & 34.2 & 20.0 & 7.5 & 32.0 & 20.0 \\
Temple Run 2    & 17.5 & 93.0 & 20.0 & 5.0 & 59.3 & 20.0 & 0.0 & 42.5 & 0.0 & 20.0 & 72.9 & 20.0 \\
\midrule
\rowcolor{gray!10}
\textbf{Overall}
  & 57.8 & 81.3 & 75.6
  & 28.1\,\ablDown{29.7} & 57.0\,\ablDown{24.3} & 44.4\,\ablDown{31.2}
  & 39.4\,\ablDown{18.4} & 64.7\,\ablDown{16.6} & 62.2\,\ablDown{13.4}
  & 40.6\,\ablDown{17.2} & 68.0\,\ablDown{13.3} & 64.4\,\ablDown{11.2} \\
\bottomrule
\end{tabular}%
}
\end{table}

\textbf{Expert-guided on-policy learning.}
We ablate the expert-guided on-policy learning stage, initializing all variants from the same SFT policy.
% and share the rollout budget and hyperparameters of \tool
To examine whether on-policy imitation (OPI) and expert-rewarded policy optimization (ERPO) are complementary, \textit{w/o ERPO} optimizes only $\ell_t^{\mathrm{OPI}}$ and \textit{w/o OPI} optimizes only $\ell_t^{\mathrm{ERPO}}$ in Eq.~\eqref{eq:rl-loss}.
To evaluate the step-criticality measure, \textit{w/o $\boldsymbol{\kappa_t}$} sets $\kappa_t\equiv1$ in Eq.~\eqref{eq:opi-weight}.
Table~\ref{tab:ablation-online-model} reports the results. 
Overall, \tool achieves 57.8\% SR, 81.3\% PR, and 75.6\% TC.
Compared with \tool, removing ERPO reduces overall SR, PR, and TC by 29.7\%, 24.3\%, and 31.2\%, while removing OPI lowers them by 18.4\%, 16.6\%, and 13.4\%, respectively.
Removing criticality weighting drops SR, PR, and TC by 17.2\%, 13.3\%, and 11.2\%.
These results show that OPI and ERPO complement each other.
OPI directly aligns the agent with expert actions on visited states, while ERPO explores sampled candidate actions by rewarding expert-equivalent choices and penalizing invalid ones.
Moreover, step-criticality weighting concentrates learning on consequential steps while relaxing supervision on fault-tolerant ones.

\begin{table}[!t]
\centering
\newcommand{\ablDown}[1]{{\tiny\textcolor{red!70!black}{$\downarrow$#1}}}
\newcommand{\ablUp}[1]{{\tiny\textcolor{green!45!black}{$\uparrow$#1}}}
\caption{Ablation on the Training Pipeline.}
\label{tab:ablation-pipeline}
\footnotesize
\setlength{\tabcolsep}{2.2pt}
\renewcommand{\arraystretch}{1.05}
\resizebox{\columnwidth}{!}{%
\begin{tabular}{l*{3}{C{2.3em}}*{3}{C{3.5em}}*{3}{>{\columncolor{blue!6}}C{3.5em}}}
\toprule
& \multicolumn{3}{c}{\makecell{\textbf{Base}\\{\color{gray}(Qwen3.5-4B)}}}
& \multicolumn{3}{c}{\makecell{\textbf{+ Action-Focused}\\{\textbf{SFT}}}}
& \multicolumn{3}{>{\columncolor{blue!6}\centering\arraybackslash}p{\dimexpr10.5em+4\tabcolsep\relax}}{\makecell{\textbf{Full Pipeline}\\{\color{gray}(\tool)}}} \\
\cmidrule(lr){2-4} \cmidrule(lr){5-7} \cmidrule(lr){8-10}
\textbf{Game} & SR & PR & TC & SR & PR & TC & SR & PR & TC \\
\midrule
Pac-Man         & 0.0 & 24.1 & 0.0 & 35.0 & 76.2 & 80.0 & 77.5 & 98.6 & 100.0 \\
Doodle Jump     & 2.5 & 25.9 & 20.0 & 32.5 & 57.8 & 40.0 & 45.0 & 82.8 & 80.0 \\
Orchard Match-3 & 0.0 & 2.5 & 0.0 & 27.5 & 35.6 & 60.0 & 45.0 & 48.8 & 100.0 \\
Chrome Dino     & 0.0 & 20.2 & 0.0 & 65.0 & 86.3 & 100.0 & 90.0 & 95.1 & 100.0 \\
Cubefield       & 0.0 & 9.0 & 0.0 & 95.0 & 96.7 & 100.0 & 82.5 & 89.9 & 100.0 \\
Edge Surf       & 2.5 & 29.3 & 20.0 & 47.5 & 90.6 & 100.0 & 60.0 & 93.9 & 60.0 \\
NS-Shaft        & 15.0 & 52.9 & 40.0 & 60.0 & 84.8 & 100.0 & 90.0 & 93.7 & 100.0 \\
Run 3           & 12.5 & 27.0 & 20.0 & 7.5 & 32.2 & 20.0 & 12.5 & 36.2 & 20.0 \\
Temple Run 2    & 10.0 & 44.8 & 20.0 & 20.0 & 50.1 & 20.0 & 17.5 & 93.0 & 20.0 \\
\midrule
\rowcolor{gray!10}
\textbf{Overall}
  & 4.7 & 26.2 & 13.3
  & 43.3\,\ablUp{38.6} & 67.8\,\ablUp{41.6} & 68.9\,\ablUp{55.6}
  & 57.8\,\ablUp{53.1} & 81.3\,\ablUp{55.1} & 75.6\,\ablUp{62.3} \\
\bottomrule
\end{tabular}%
}
\end{table}

\textbf{Training pipeline.}
To quantify the contribution of each training stage, we evaluate three models along the training pipeline of \tool:
the untrained base model Qwen3.5-4B,
the policy after action-focused SFT on offline expert trajectories,
and the full \tool pipeline, which further applies expert-guided on-policy learning initialized from the SFT policy.
Table~\ref{tab:ablation-pipeline} reports the results.
The untrained base model struggles across tasks, attaining only 4.7\% SR, 26.2\% PR, and 13.3\% TC.
Adding action-focused SFT substantially boosts SR, PR, and TC by 38.6\%, 41.6\%, and 55.6\% to 43.3\%, 67.8\%, and 68.9\%, respectively.
Further applying expert-guided on-policy learning raises performance to 57.8\% SR, 81.3\% PR, and 75.6\% TC, surpassing the base model by 53.1\%, 55.1\%, and 62.3\%.
These results show that action-focused SFT provides a strong cold start by grounding multimodal observations into accurate actions, while expert-guided on-policy learning enables the agent to recover and generalize to game states unseen during SFT.

% !TeX root = ../main.tex

\section{Discussion}\label{discussion}

\textbf{Conclusion.}
\rev{We present \tool, a three-stage framework for training a unified visual game agent across diverse games, leveraging programmatic experts as offline demonstrators and online step-level judges.
\tool first synthesizes verified expert trajectories, then cold-starts the agent via action-focused SFT, and finally optimizes it over truncated rollouts through criticality-weighted on-policy imitation and ERPO.
Our experiments demonstrate that our agent significantly outperforms state-of-the-art baselines.}

\textbf{Limitations.}
\rev{\looseness=-1 First, \tool relies on programmatic experts that read internal game states during training. Yet the agent requires no such access at inference, acting solely from screenshots and device states.
Second, imitating deterministic experts may constrain behavioral diversity. We alleviate this issue by varying episode seeds and relaxing supervision on non-urgent steps via criticality weighting.
Future work can automate expert construction with code-generation models and explore zero-shot generalization to unseen games.}

\section{LLM Usage Statement}\label{llm-usage-statement}

For manuscript preparation, we used LLMs solely to proofread and polish the text.
As part of our method, a VLM (i.e., Qwen3.5-397B-A17B) is used to generate step annotations for the expert trajectories (Appendix~\ref{sec:annotation-prompt}).
All research ideas, experimental designs, and final content were conceived, verified, and approved by the authors, who take full responsibility for this paper.

\bibliographystyle{plainnat}
\bibliography{src/references}

\appendix
% !TeX root = ../main.tex

\lstdefinestyle{prompt}{basicstyle=\ttfamily\footnotesize,breaklines=true,columns=fullflexible,keepspaces=true,showstringspaces=false,frame=single}

\section{Game Descriptions}\label{sec:game-descriptions}

\begin{table}[t]
\begin{revblock}
\centering
\caption{\rev{\justifying Games and Genres. The nine games used in this work.}}\label{tab:game-descriptions}
\footnotesize
\setlength{\tabcolsep}{5pt}
\renewcommand{\arraystretch}{1.05}
\begin{tabularx}{\linewidth}{@{}l l >{\raggedright\arraybackslash}X@{}}
\toprule
\textbf{Genre} & \textbf{Game} & \textbf{Gameplay} \\
\midrule
Arcade & Pac-Man & Clear a maze of pellets while evading the ghosts \\
\midrule
Platformer & Doodle Jump & Bounce upward across moving and breakable platforms \\
\midrule
Puzzle & Orchard Match-3 & Swap adjacent fruits to match three or more and reach the score goal \\
\midrule
Runner & Chrome Dino & Jump over cacti and duck under birds at rising speed \\
\addlinespace[2pt]
 & Cubefield & Steer a ship through a 3D field of cubes without crashing \\
\addlinespace[2pt]
 & Edge Surf & Surf past obstacles with three lives, collecting boosts and shields \\
\addlinespace[2pt]
 & NS-Shaft & Descend a shaft by landing on shifting platforms and avoiding hazards \\
\addlinespace[2pt]
 & Run 3 & Run and jump across gaps in a rotating space tunnel \\
\addlinespace[2pt]
 & Temple Run 2 & Switch lanes, turn, jump, and slide in a high-speed 3D run \\
\bottomrule
\end{tabularx}
\end{revblock}
\end{table}

This section describes the \rev{nine browser games} used in this work, grouped into \rev{four} genres as summarized in Table~\ref{tab:game-descriptions}.
Each game is played through keyboard and mouse inputs, and each game is associated with five tasks that specify concrete gameplay objectives.
\rev{The five tasks of a game increase in difficulty, such as reaching a target score, a target distance, or a number of cleared levels within a step budget, giving 45 tasks in total.
Table~\ref{tab:game-tasks} lists the success condition and the step budget of every task.}

\textbf{Arcade.} Arcade games require fast closed loop control with frequent reactions to moving entities.
In \texttt{Pac-Man}, the player guides Pac-Man through a maze to eat all pellets and avoid the ghosts, which become edible for a short time after Pac-Man eats a power pellet.

\textbf{Platformer.} Platformer games require precise movement and jumping across structured terrain while avoiding hazards.
In \texttt{Doodle Jump}, the character keeps bouncing upward and the player steers it left or right to land on higher platforms, some of which move, break, or vanish after one use.

\textbf{Puzzle.} Puzzle games emphasize reasoning over discrete states and planning over long horizons.
\rev{In \texttt{Orchard Match-3}, the player drags a fruit onto an adjacent cell to swap the two, and every line of three or more identical fruits is cleared for points.
Special pieces such as wild fruits, bombs, and rainbows clear additional fruits, and limited tools such as a hammer can be clicked and applied to a chosen cell.
Each level is cleared once its score reaches the goal of that level within a limited number of moves.}

\textbf{Runner.} Runner games advance the scene automatically and require timely reactions to approaching obstacles.
\texttt{Chrome Dino} is the endless runner of the Chrome browser, where the dinosaur jumps over cacti and ducks under birds flying low as its speed increases.
\texttt{Cubefield} places the player in a triangular ship that flies forward through a 3D field of cubes, and the player steers left or right because hitting any cube ends the run.
In \texttt{Edge Surf}, the player steers a surfer to avoid obstacles and collect items, where boosts grant temporary speed, shields grant temporary invincibility, and each collision costs one of three lives.
In \texttt{NS-Shaft}, the character descends through a shaft of shifting platforms, and the player moves it left or right to land safely, since touching hazards reduces its health and falling to the bottom ends the game.
\texttt{Run 3} takes place in a tunnel in space, where the player strafes and jumps to cross gaps, and moving onto a side wall rotates the tunnel so that the wall becomes the new floor.
\texttt{Temple Run 2} is a 3D endless runner at high speed, in which the player switches lanes, turns at intersections, jumps over obstacles, and slides under them, and any missed action ends the run.

\begin{table*}[t]
\begin{revblock}
\centering
\newcommand{\task}[2]{\makecell{#1\\[-1.5pt]{\scriptsize\color{black!50}#2}}}
\caption{\rev{\justifying Tasks of Each Game. Each cell gives the threshold that the objective of the game must reach for the task to succeed, and the gray number below it gives the step budget of the task. Every task must be completed in a single run, subject to the additional constraint of its game.}}\label{tab:game-tasks}
\footnotesize
\setlength{\tabcolsep}{4.5pt}
\renewcommand{\arraystretch}{1.1}
\begin{tabularx}{\textwidth}{@{}l l l *{5}{c} >{\raggedright\arraybackslash}X@{}}
\toprule
 & & & \multicolumn{5}{c}{\textbf{Task}} & \\
\cmidrule(lr){4-8}
\textbf{Genre} & \textbf{Game} & \textbf{Objective} & \textbf{01} & \textbf{02} & \textbf{03} & \textbf{04} & \textbf{05} & \textbf{Additional Constraint} \\
\midrule
Arcade & Pac-Man & Score & \task{600}{1,500} & \task{1,000}{1,500} & \task{1,500}{3,000} & \task{2,000}{3,000} & \task{2,500}{3,000} & -- \\
\midrule
Platformer & Doodle Jump & Score & \task{200}{500} & \task{500}{500} & \task{1,000}{1,200} & \task{1,500}{1,800} & \task{2,000}{2,400} & -- \\
\midrule
Puzzle & Orchard Match-3 & Total score / levels & \task{300}{500} & \task{1 level}{500} & \task{2 levels}{1,000} & \task{3 levels}{1,000} & \task{4 levels}{1,000} & Levels 1--4 need 1,000 / 1,450 / 1,900 / 2,550 points \\
\midrule
Runner & Chrome Dino & Score & \task{100}{500} & \task{200}{540} & \task{400}{960} & \task{600}{1,320} & \task{800}{1,650} & -- \\
\addlinespace[2pt]
 & Cubefield & Score & \task{70K}{850} & \task{90K}{1,100} & \task{115K}{2,700} & \task{160K}{3,700} & \task{220K}{5,000} & No crash; task 05 also requires level 2 \\
\addlinespace[2pt]
 & Edge Surf & Score & \task{200}{500} & \task{500}{500} & \task{800}{1,600} & \task{1,200}{2,400} & \task{2,000}{4,000} & Task 01 loses no life \\
\addlinespace[2pt]
 & NS-Shaft & Survival time (s) & \task{6}{500} & \task{10}{500} & \task{15}{500} & \task{20}{540} & \task{25}{660} & -- \\
\addlinespace[2pt]
 & Run 3 & Distance & \task{50}{500} & \task{150}{500} & \task{300}{1,600} & \task{450}{3,000} & \task{600}{3,000} & Infinite mode; no fall \\
\addlinespace[2pt]
 & Temple Run 2 & Score & \task{2,000}{500} & \task{4,000}{3,000} & \task{6,000}{4,000} & \task{8,000}{4,000} & \task{10,000}{4,000} & No stumble, death, or resurrection \\
\bottomrule
\end{tabularx}
\end{revblock}
\end{table*}

\rev{
\section{Expert Models}\label{sec:expert-models}

This section describes the expert model of each game introduced in Section~\ref{sec:experts}.
All experts share one interface.
At each 50\,ms control step, an expert receives the structured game state and the current input device state, and returns the complete set of keys to hold after the step, together with the keys to tap and the mouse drags to perform.
An expert keeps a memory across steps, such as the target it is currently pursuing, and updates this memory according to the action that was actually executed.
An expert raises an error instead of falling back to waiting when it cannot interpret the game state, so that a misread state surfaces as a failure during refinement rather than as a silent wait in the trajectory.

\textbf{Pac-Man.}
The expert reads the tile grid of the maze, including walls, pellets, and power pellets, together with the positions and states of the ghosts.
Each decision selects a neighboring tile to head for, and the expert keeps its move until Pac-Man has entered that tile, so that a queued turn is not canceled by replanning at intermediate positions.
The danger of each tile is measured along the corridors from every live ghost.
When a ghost closes in within four tiles, the expert escapes toward the neighbor whose short lookahead keeps the largest clearance, preferring power pellets and entering dead ends only as a last resort.
An edible ghost within reach is chased, and otherwise the expert heads for the remaining pellets.

\textbf{Doodle Jump.}
During each flight, the expert considers every visible platform that the falling character can still reach, computing the descending crossing of the platform top under gravity and the sideways distance reachable by holding an arrow key until then.
It keeps the target chosen for the current flight while it remains reachable, and otherwise selects the soonest and nearest platform, preferring springs and stable platforms and avoiding the platform it just left when a higher one is reachable.
The expert then steers so that coasting with the arrow keys released ends within the landing band of the target, which avoids rapid switching between the two keys.

\textbf{Orchard Match-3.}
The expert applies the match-three rules, including wild fruits, bombs, and rainbows, to the visible board and ranks all legal swaps by their estimated score.
It uses a hammer or a bomb instead of a swap when the gain per required click exceeds that of the best swap, or when few moves remain.
Refills of the board are unknown before they appear and are therefore never counted in the estimates.

\textbf{Chrome Dino.}
The expert models the jump of the dinosaur with the physics and hitboxes of the game.
A takeoff clears an obstacle when it starts between about 4 and 30 frames of travel before the obstacle, and the expert taps in the middle of this window so that a deviation of one control step still clears it.
A closely following obstacle moves the takeoff earlier so that the dinosaur lands in time for the next jump.
Birds flying at head height or above are avoided by holding the duck key until they have passed.

\textbf{Cubefield.}
The ship banks while an arrow key is held and levels out by itself once the key is released, and its bank determines its lateral speed.
The expert performs a beam search over left, right, and no steering for 24 control steps, i.e., 1.2 seconds.
A plan must stay inside the walls outlined by the visible cubes and never touch a cube, and it is scored by its clearances, its steering effort, and how well it stays in the middle of the passage.
The expert executes the first action of the best plan.

\textbf{Edge Surf.}
The base policy blocks a band of lanes for each obstacle ahead, assigns each lane a cost from its overlap with these bands weighted by their distance, and taps toward the cheapest reachable lane when it is clearly better than staying.
After refinement, a frame-level simulator reproduces the heading and speed of the surfer, the hitboxes of all objects, the zigzag motion of the other surfers, and the pursuit of the kraken.
A collision shield runs the base policy 1.2 seconds ahead in this simulator and replaces its action only when a crash is predicted.
When a kraken pursues the surfer, the expert searches for riding patterns that hold one direction and then switch to the other, and selects the pattern that makes the kraken hit an obstacle while the surfer stays clear.

\textbf{NS-Shaft.}
Each descent is planned from the visible scene.
The character stays on its current floor until a safe floor below is completely visible in the screenshot, walks off the edge on the side of that floor, steers until it is above the floor, and then releases the keys to land on it.
Each candidate descent is simulated with the native physics of the game, so that the chosen walk-off actually ends on the intended floor.
Floors still below the bottom edge of the screenshot are never a reason to move.

\textbf{Run 3.}
The game state lists the collidable tiles near the runner in its gravity frame, and the expert merges tiles that share a plane so that the tunnel becomes a polygon with holes.
At every airborne step, the expert replans by simulating candidate steering pulses and jump durations with a motion model fitted to recorded browser traces, until the runner touches a tile, falls through a hole, or hits an obstacle.
Each candidate landing is scored by its distance from the tile boundaries, and the expert executes the first action of the best candidate.

\textbf{Temple Run 2.}
The runner tilts continuously instead of switching between fixed lanes, a turn only registers near the middle of a turn piece, and repeated jump taps queue additional jumps.
The expert therefore issues exactly one tap for each feature of the track, such as a turn, a junction, a jump, or a slide, inside the valid timing window of that feature, and tilts toward the side of narrow ledges early enough to reach it in time.
}

\subsection{\texorpdfstring{\rev{Iterative Refinement}}{Iterative Refinement}}\label{sec:expert-refinement}

\rev{Algorithm~\ref{alg:expert-refinement} summarizes the refinement of the expert of one game, introduced in Section~\ref{sec:experts}.
In round $k$, we roll out the current expert $E^{(k)}$ for $M$ episodes on each task, each episode with a different seed, and the task evaluator labels every episode as a success or a failure.
For each failed episode, we inspect the recorded screenshots, game states, and actions to locate the earliest step after which the failure can no longer be avoided, and we attribute the failure to one of five causes, i.e., a misread game state, inaccurate dynamics in the forward model, an entity or hazard that the expert does not model, an action issued outside its valid timing window, and a planning horizon that is too short to avoid a hazard.
We revise the expert modules responsible for the most frequent causes and add the failure situations to the regression tests of the expert.
Before the revised expert $E^{(k+1)}$ replaces $E^{(k)}$, we replay the trajectories recorded in earlier rounds and check that $E^{(k+1)}$ reproduces their actions, except at the steps targeted by the revision.
The refinement of a game ends once its expert reaches a success rate of at least $\eta=50\%$ on every task.
Table~\ref{tab:data-source} reports the task coverage and success rate of every refined expert.}

% Floats reset the text color, so revblock goes inside the float.
\begin{algorithm}[!htbp]
\begin{revblock}
\caption{Iterative expert refinement for one game.}\label{alg:expert-refinement}
\KwIn{Tasks $\mathcal{J}$ of the game, initial expert $E^{(0)}$, rollouts per task $M$, target success rate $\eta$}
\KwOut{Refined expert $E^{(k)}$}
$k \leftarrow 0$\;
\While{true}{
  \ForEach{task $j \in \mathcal{J}$}{
    Roll out $E^{(k)}$ for $M$ episodes on task $j$ with different seeds\;
    $p^{(k)}_j \leftarrow$ fraction of episodes that the task evaluator verifies as successful\;
  }
  \lIf{$\min_{j \in \mathcal{J}} p^{(k)}_j \geq \eta$}{\Return{$E^{(k)}$}}
  \ForEach{failed episode $\tau$}{
    Locate the earliest step after which the failure of $\tau$ is unavoidable\;
    Attribute the failure to a cause $c(\tau)$\;
  }
  Revise the modules of $E^{(k)}$ responsible for the most frequent causes to obtain $E^{(k+1)}$\;
  Add the failure situations to the regression tests of the expert\;
  Replay the trajectories of earlier rounds and verify that $E^{(k+1)}$ reproduces their actions except at the revised steps\;
  $k \leftarrow k + 1$\;
}
\end{revblock}
\end{algorithm}

\begin{revblock}
\section{Agent Prompts}\label{sec:agent-prompts}

This section presents the prompts through which \tool interacts with the games, as formulated in Section~\ref{sec:preliminaries}.
Each decision step is formulated as a single conversational turn consisting of a system prompt, an instruction prompt, and the model response.
The system prompt carries the task context $D$ together with the specification of the action space and the response format, and remains fixed throughout an episode.
The instruction prompt is rebuilt at every step from the history $h_t^K$, the input device state $s_t$, and the current screenshot $v_t$.
The response contains the three target segments in Equation~\ref{eq:agent-generation}.
The supervised fine-tuning samples, the reinforcement learning rollouts, and the online evaluation requests are built with the same prompt builders, so that \tool receives identical prompt text during training and inference.
All baselines in Section~\ref{sec:exp-setup} are evaluated with the same prompts.

\subsection{System Prompt}\label{sec:agent-system-prompt}

The system prompt is instantiated from the following template.

\begin{lstlisting}[style=prompt]
# GUI Game Agent

You are an autonomous agent that plays a GUI game by driving its keyboard and mouse, exactly as a human player would. At each step you receive the current screenshot and the input state that reports the cursor position and which keys and buttons are physically held right now. Your job is to decide the next action that best advances the task objective.

## Game

{game}

## Task

{task_prompt}

## Game Rules and Controls

{guidance}

{action_reference}

{response_format}
\end{lstlisting}

The placeholders \texttt{\{game\}} and \texttt{\{task\_prompt\}} are filled with the game identifier and the natural language objective of the current task, respectively.
The placeholder \texttt{\{guidance\}} is filled with a game-specific guide that describes the game objective, the game rules, and the player controls, and concludes with the legal inputs of the game.
The placeholders \texttt{\{action\_reference\}} and \texttt{\{response\_format\}} are shared across all games and are presented in Appendix~\ref{sec:unified-action} and Appendix~\ref{sec:agent-response-format}, respectively.
For example, the system prompt of the first \texttt{Doodle Jump} task in Table~\ref{tab:game-tasks} instantiates the game-dependent placeholders as follows.

\begin{lstlisting}[style=prompt]
## Game

doodle-jump

## Task

Build a stable early climb and finish with a score of at least 200.

## Game Rules and Controls

#### Doodle Jump

You are playing **Doodle Jump**, a vertical platformer in which Doodle bounces
automatically upward through an endlessly scrolling stack of platforms.

##### Game Objective

- Keep ascending and maximize the vertical-distance score.
- A landing on a higher platform continues the run; falling below the screen
  ends it.

##### Game Rules

- Doodle has no manual jump button. Landing on a platform automatically starts
  the next bounce, while gravity brings Doodle back down.
- Green platforms are stable and can be reused. Blue platforms move
  horizontally, so their future position matters. Red platforms are breakable
  and give way after being used; white platforms vanish after one use.
- The screen scrolls upward as the score grows. Platforms are generated ahead
  of the camera, and their horizontal spacing becomes more demanding over
  time.
- The left and right screen edges wrap: leaving one side re-enters from the
  opposite side. This can rescue a landing, but it also makes a long hold wrap
  farther than expected.
- There are no attacks or inventory items in this recreation. The only threat
  is missing a platform or colliding with an unsafe platform state.

##### Doodle Controls

You control Doodle's horizontal drift while vertical motion is automatic.
Horizontal velocity has acceleration and braking, so release a direction early
when lining up a narrow platform.

ACTION SPACE (ONLY LEGAL ACTIONS):

- Wait: Continue the current bounce and let horizontal velocity slow down.
- ArrowLeft: Accelerate left while held.
- ArrowRight: Accelerate right while held.
- Space: Start a new run from the title/game-over screen or reset after death.
\end{lstlisting}

The game guides are the same as those used for trajectory collection and annotation (Appendix~\ref{sec:annotation-system-prompt}), except that the guide of \texttt{Orchard Match-3} writes its two example swap actions in the action representation of Appendix~\ref{sec:unified-action} rather than as JSON objects.

\subsection{Action Space}\label{sec:unified-action}

The action reference specifies the action space $\mathcal{A}$ of Section~\ref{sec:preliminaries}, which is shared across all games and written as plain text.
An action is either a standalone word, i.e., \texttt{noop} for a waiting step or \texttt{complete} for ending the episode, or one or more operations separated by semicolons.
Keyboard keys and mouse buttons share the \texttt{release}, \texttt{hold}, and \texttt{tap} operations, whereas \texttt{move} and \texttt{drag} take points whose coordinates are integers from 0 to 1000, normalized by the width and height of the screenshot.
The reference further specifies the persistence of held inputs and a canonical order that gives every action exactly one valid serialization, and concludes with examples of complete actions.

\begin{lstlisting}[style=prompt]
## Available Actions

Every step you emit exactly one action wrapped in `<act> ... </act>`.

### Action Shape

An action is either a single word or one or more operations separated by `; `.

| Action | Meaning |
| --- | --- |
| `noop` | Advance one step without changing any held input. Must stand alone. |
| `complete` | End the episode. Use it only once the task objective is met. Must stand alone. |

### Keyboard Operations

| Syntax | Meaning |
| --- | --- |
| `release KEY ...` | Release the listed keys. |
| `hold KEY ...` | Press the listed keys down and keep them held across later steps until released. |
| `type "TEXT"` | Type the text character by character. |
| `tap KEY ...` | Tap the listed keys once (down, then up); does not change the held set. |

Key names follow the game controls, e.g. `w`, `1`, `Space`, `Enter`, `Escape`, `Shift`, `ArrowUp`. Arrow keys are always written `ArrowLeft`, `ArrowRight`, `ArrowUp` and `ArrowDown`.

### Mouse Operations

| Syntax | Meaning |
| --- | --- |
| `move [x, y]` | Move the cursor to the point. |
| `release BUTTON ...` | Release the listed mouse buttons. |
| `hold BUTTON ...` | Press the listed mouse buttons down and keep them held across later steps until released. |
| `tap BUTTON ...` | Click the listed mouse buttons once (down, then up); does not change the held set. |
| `drag [x1, y1] [x2, y2]` | Press the left button at the first point, glide to the second point and release, all within this step. |

Mouse buttons are `left`, `middle` and `right`. They are reserved names and never mean keyboard keys.

Mouse buttons carry no point: they always act at the current cursor position. To press or click somewhere else, put `move [x, y]` in the same action; the cursor moves before buttons go down, so `move [x, y]; tap left` clicks at `[x, y]`. `drag` positions the cursor by itself, so it cannot be combined with `move`, `hold left` or `tap left`.

### Combining Keyboard and Mouse

Keyboard keys and mouse buttons share the same `release`, `hold` and `tap` operations. When one step uses both, list them after a single operation, keys first: `tap Space left` taps Space and clicks the left button. A name may appear in only one of `release`, `hold` or `tap` within one action.

### Coordinates

A point `[x, y]` is a pair of integers from 0 to 1000 relative to the current screenshot, with `[0, 0]` at the top-left and `[1000, 1000]` at the bottom-right.

- `x` runs from the left edge (0) to the right edge (1000).
- `y` runs from the top edge (0) to the bottom edge (1000).
- `[0, 0]` is the top-left corner, `[1000, 1000]` is the bottom-right corner and `[500, 500]` is the center of the image.

### Input State

Each step reports the current input state as JSON. `cursor` is the cursor position on the same 0 to 1000 scale, `keys_held` lists the keys held right now and `mouse_held` lists the mouse buttons held right now.

### Held State and Timing

- A hold stays down across later steps until you release it. Inputs you do not mention keep their current state.
- A tap is a one-off event and never changes what is held.
- To hold something for a while, hold it once, wait for as many steps as you want, then release it.
- Within one step the operations take effect in the order they are written: releases, cursor movement, new holds, typed text, taps, then the drag.

### Canonical Order

Every action has exactly one valid serialization. Order it by these rules.

1. Operations come in this order: `release`, `move`, `hold`, `type`, `tap`, `drag`.
2. Each operation appears at most once.
3. The names after `release`, `hold` or `tap` are listed once each: keyboard keys first in ASCII order (digits, then capitalized names such as `ArrowUp` `Shift` `Space`, then lowercase letters), then mouse buttons in the order `left`, `middle`, `right`.

### Examples

| Intent | Action |
| --- | --- |
| Wait one step, keeping every held input | `<act>noop</act>` |
| Switch from running left to running right | `<act>release ArrowLeft; hold ArrowRight</act>` |
| Start running forward while jumping once | `<act>hold Shift w; tap Space</act>` |
| Click a button | `<act>move [842, 512]; tap left</act>` |
| Flag a cell with the right button | `<act>move [498, 181]; tap right</act>` |
| Click again where the cursor already is | `<act>tap left</act>` |
| Press the left button on an item and keep it held | `<act>move [310, 640]; hold left</act>` |
| Release the held left button where the cursor is | `<act>release left</act>` |
| Drag an item across the board | `<act>drag [200, 300] [700, 300]</act>` |
| Type a word into a text field | `<act>type "crane"</act>` |
| Combine keyboard and mouse in one step | `<act>move [516, 275]; hold Shift d; tap Space left</act>` |
| Finish once the objective is met | `<act>complete</act>` |
\end{lstlisting}

The expert trajectories record each action as a JSON object with coordinates in screenshot pixels (Appendix~\ref{sec:annotation-system-prompt}), and we convert every recorded action into this representation before training.
The \texttt{hold}, \texttt{tap}, and \texttt{release} lists of the keyboard and mouse channels become the operations of the same names, with keys listed before buttons.
The cursor target of an absolute movement and the endpoints of a drag are mapped from a pixel position $(x, y)$ to the point $[\mathrm{round}(1000x/W), \mathrm{round}(1000y/H)]$, where $W \times H$ is the screenshot size.
Typed text becomes a \texttt{type} operation, an empty action becomes \texttt{noop}, and a completion becomes \texttt{complete}.
Relative cursor movement and scrolling are not required by any of the nine games and are therefore not part of the action space.
At inference, the points of a predicted action are mapped back to pixels of the current viewport before the action is executed.

\subsection{Instruction Prompt}\label{sec:agent-instruction-prompt}

The instruction prompt forms the user message of each decision step.
Together with the task context $D$ in the system prompt, it realizes the model input $x_t$ in Equation~\ref{eq:agent-generation}.
It consists of a recent history section, which presents the history $h_t^K$, followed by a decision section.
The history section begins with a heading and a short description of its role, followed by one entry per step in chronological order.
The entry of step $i$ consists of a header, the screenshot $v_i$, and the action $a_i$ in its canonical form.
The rationale and goal of earlier steps are not included in the history.
Entries are numbered by their position in the window rather than by their step in the episode, and the current screen carries no step number, so that the model cannot infer task progress from the step count.
The decision section then reports the input device state $s_t$ and precedes the current screenshot $v_t$.
The complete template is shown below, where \texttt{[screenshot of entry k]} and \texttt{[current screenshot]} mark the positions at which the corresponding screenshots are inserted.

\begin{lstlisting}[style=prompt]
# Recent History

This section holds the most recent steps leading up to now. Each entry shows the screen observed at that step, before its action took effect, followed by the action you took at that step; compare consecutive screens with the current one to see what just happened.
Entries are listed oldest first and numbered from 1, so the highest number is the step right before the current screen.

## Step {k}

[screenshot of entry k]

{action of entry k}

...(one entry per step in the history window)

# Decision

Choose the next action that best advances the task objective.

## Current Input State

```json
{input_state}
```

## Current Screen

[current screenshot]
\end{lstlisting}

With the history window of $K=1$ in Section~\ref{sec:preliminaries}, the history section holds a single entry, i.e., the step right before the current screen.
In the training samples, steps without a stored screenshot (Appendix~\ref{sec:annotation-request}) are skipped, so this entry is the most recent step whose screenshot is stored.
At the first step of an episode, the history window is empty, and its entries are replaced by the sentence ``No steps have been recorded yet.''
The placeholder \texttt{\{input\_state\}} is filled with a JSON object that reports the cursor position on the same 0 to 1000 scale as the actions, the held keys, and the held mouse buttons.
For example, the following instruction prompt is taken from a successful trajectory of the \texttt{Doodle Jump} task above, at a step where Doodle drifts to the right while \texttt{ArrowRight} is held and the agent waited at the previous step.

\begin{lstlisting}[style=prompt]
# Recent History

This section holds the most recent steps leading up to now. Each entry shows the screen observed at that step, before its action took effect, followed by the action you took at that step; compare consecutive screens with the current one to see what just happened.
Entries are listed oldest first and numbered from 1, so the highest number is the step right before the current screen.

## Step 1

[screenshot of the previous step]

<act>noop</act>

# Decision

Choose the next action that best advances the task objective.

## Current Input State

```json
{"cursor": [500, 500], "keys_held": ["ArrowRight"], "mouse_held": []}
```

## Current Screen

[current screenshot]
\end{lstlisting}

\subsection{Response Format}\label{sec:agent-response-format}

The system prompt concludes with the following response format, which requires the model to generate the action, rationale, and goal in the order of Equation~\ref{eq:agent-generation}.

\begin{lstlisting}[style=prompt]
## Response Format

Your reply has three sections. `<act>` carries the action, `<think>` holds a brief analysis of the observation and history, and `<goal>` states the current objective. Emit them in this exact order, with no JSON, headings or Markdown fences:

```text
<act>One action.</act>
<think>Brief analysis of the observation and history.</think>
<goal>One immediate objective.</goal>
```
\end{lstlisting}

The \texttt{<act>}, \texttt{<think>}, and \texttt{<goal>} segments carry the action $a_t$, the rationale $r_t$, and the goal $g_t$, respectively.
During training, the target response contains all three segments and is followed by an end-of-turn token, and the rationale and goal are HTML-escaped so that their text cannot contain the tags of the response format.
At inference, decoding stops at \texttt{</act>}, and the action inside the \texttt{<act>} block is parsed and executed as described in Appendix~\ref{sec:unified-action}.
The following two target responses are taken from the same successful trajectory of the \texttt{Doodle Jump} task as the instruction prompt above, with the rationale and goal annotated as described in Appendix~\ref{sec:annotation-prompt}.
The first response answers that instruction prompt, where the agent releases \texttt{ArrowRight} and holds \texttt{ArrowLeft} within one step to reverse the drift of Doodle.
In the second response, the agent emits \texttt{complete} after the score reaches the target of 200.
Line breaks between segments are added for readability.

\begin{lstlisting}[style=prompt]
<act>release ArrowRight; hold ArrowLeft</act>
<think>With ArrowRight held, Doodle is drifting away from the nearest landing spot on the left. I need to switch inputs to ArrowLeft to correct the trajectory and secure the next bounce.</think>
<goal>Reverse horizontal direction to target the left platform</goal>
\end{lstlisting}

\begin{lstlisting}[style=prompt]
<act>complete</act>
<think>The score display shows 201, which meets the objective of reaching at least 200 points, so I declare the task complete.</think>
<goal>Finish the episode successfully</goal>
\end{lstlisting}
\end{revblock}

\section{Trajectory Annotation Prompt}\label{sec:annotation-prompt}

This section presents the prompt used to annotate each trajectory step with a rationale, and a goal, as described in Section~\ref{sec:experts}.
Annotation operates on the recorded trajectories before their actions are converted into the action representation of Appendix~\ref{sec:unified-action}.
In these recordings, each action is stored as a JSON object that groups device operations by channel, with coordinates given in screenshot pixels.
The annotation prompt therefore describes actions in this JSON form.
Each annotation request consists of a system prompt and a user message that presents the step to be annotated.

\subsection{Annotation System Prompt}\label{sec:annotation-system-prompt}

The system prompt consists of \rev{a session overview, a game prompt, and annotation instructions}.
The game prompt is the agent system prompt under which the trajectories are recorded, which is also the prompt of the VLM agent pipeline in Section~\ref{sec:experts}.
It describes how to play the current game and how to express actions, and is organized into seven sections.
\begin{itemize}[leftmargin=*,itemsep=0pt,topsep=2pt]
    \item \textbf{Role} describes the model as a computer-use agent that plays games. It asks the model to inspect the play area, the heads-up display, and any pop-ups in the screenshot, and to combine them with the reported input state and recent actions when choosing the next action.
    \item \textbf{Coordinate System} states the screenshot resolution of $1280\times800$ pixels and defines absolute positions and relative displacements in pixels, measured from the upper left corner.
    \item \textbf{Action Space} defines an action as a JSON object with four optional device channels, i.e., \texttt{keyboard}, \texttt{mouse}, \texttt{move}, and \texttt{scroll}, and a \texttt{complete} flag. Held inputs follow the same persistence rules as in Appendix~\ref{sec:unified-action}, and each recorded action is converted into that representation as described there, where an empty object \texttt{\{\}} becomes \texttt{noop}.
    \item \textbf{Response Format} specifies the JSON reply of a decision step and defines the \texttt{goal}, \texttt{thought}, and \texttt{expected\_outcome} fields, which the annotation instructions refer to. We reproduce it below.
    \item \textbf{Examples} provides seven worked replies, covering point-and-click interaction, multi-channel aiming and firing, waiting, holding one key while tapping another, holding a button across steps, firing while aiming with a held button, and task completion.
    \item \textbf{Game-specific Guidance} contains the same game guide as the agent system prompt, e.g., the \texttt{Doodle Jump} guide in Appendix~\ref{sec:agent-system-prompt}.
    \item \textbf{Task Objective} states the objective of the current task and the completion rules, as reproduced below.
\end{itemize}

\begin{lstlisting}[style=prompt]
### Response Format

Your entire response must consist of exactly one valid JSON object representing a single `Action`.

Output nothing except the JSON object. Do not include explanatory prose, Markdown code fences, comments, headings, or text before or after the object.

Every field is optional unless required by a condition described below. Include only the fields needed for the current step.
Do not output placeholder values such as `"..."`, literal `OR` expressions, comments, or trailing commas. Do not output any duration, time or wait field: the game time per step is fixed by the environment.

The `Action` may contain the following fields:

```text
{
  "keyboard": <keyboard command>,
  "mouse": <mouse command>,
  "move": <cursor movement command>,
  "scroll": <scroll command>,
  "complete": <boolean>,
  "reason": <string>,
  "goal": <string>,
  "thought": <string>,
  "expected_outcome": <string>
}
```

#### Allowed Field Forms

- `keyboard` is an object with any subset of these fields:

  - `"hold": ["w", ...]`
  - `"tap": ["w", ...]`
  - `"release": ["w", ...]`
  - `"text": "..."`

  A key must not appear in more than one of `hold`, `tap` and `release`.

- `mouse` is an object with any subset of these fields:

  - `"hold": ["left", ...]`
  - `"tap": ["left", ...]`
  - `"release": ["left", ...]`
  - `"drag": {"x1": 0, "y1": 0, "x2": 0, "y2": 0}`

  A button must not appear in more than one of `hold`, `tap` and `release`. `drag` must not be combined with `"left"` in `hold` or `tap`, nor with a `move` channel.

- `move` must use exactly one of the following forms:

  - `{"mode": "absolute", "x": 0, "y": 0}`
  - `{"mode": "relative", "dx": 0, "dy": 0}`

- `scroll` must use the following form:

  - `{"dx": 0, "dy": 0}`

- `complete` indicates whether the task has finished:

  - `"complete": true`
  - `"complete": false`

- `reason` briefly explains why the task ended. Include it only when `complete` is `true`. When `complete` is `true`, `reason` is required and must contain exactly one concise sentence.

- `goal` states the short, high-level objective served by the current action.

- `thought` gives a concise rationale for choosing the current action based on the visible game state, the reported input state and recent action history. Keep it brief and do not provide lengthy step-by-step reasoning.

- `expected_outcome` briefly describes the immediate result expected from the action.

#### Completion Rules

If the task is still in progress:

- Omit `reason`.
- Either omit `complete` or set it to `false`.

If the task is finished:

- Set `complete` to `true`.
- Include a one-sentence `reason`.
- Omit all device channels because they are ignored after completion. Every held input is released automatically when the episode ends.

Preferred completion format:

`{"complete": true, "reason": "The objective has been successfully completed."}`
\end{lstlisting}

\begin{lstlisting}[style=prompt]
### Task Objective

You must accomplish the following task objective:

{task_prompt}

#### How this task ends

- The episode ends ONLY when YOU reply with `"complete": true`, or when the step budget is exhausted.
- Reply with `"complete": true` as soon as the objective above is satisfied. Do NOT declare completion before it is met.
\end{lstlisting}

\rev{A session overview is placed before the game prompt, and annotation instructions are appended after it.
The overview tells the model that it is the player of an episode that has already been played to the end, and that the game prompt reproduces the instructions under which the episode was played.}

\begin{revblock}
\begin{lstlisting}[style=prompt]
### Session Overview

You are the player of a game episode that has already been played to the end. In this session you look back at one step of that episode and write down what was in your mind when you took its action.

The sections between this overview and the final `Annotation Mode` section reproduce the instructions you played under. Read them to recall the game, the task and what every action means. They are background for this session: where they tell you to choose an action or to reply with an `Action` object, they describe how you played, not what you do now. What you do now is described in `Annotation Mode`.
\end{lstlisting}
\end{revblock}

The \rev{annotation instructions} change the role of the model from choosing an action to explaining an action that the player has already executed.
They also require the model to write from the perspective of the player and to phrase the expected outcome as a prediction, without revealing that the subsequent screenshot has been observed.
\rev{They further forbid the model from stating anything that only the subsequent screenshot reveals, so that each annotation contains only information available at the moment of the decision.}

\begin{revblock}
\begin{lstlisting}[style=prompt]
### Annotation Mode

The user message first replays the step exactly as you saw it while playing: your action history, the current input state and the current game screen. It still speaks of "right now" and asks for your next action, because it is the original request of that step. Two sections follow that you did not have at the time:

- `Executed Action` is the action you took on the current game screen, written as the `Action` JSON described in `Action Space`.
- `Game Screen After The Action` is the game screen one step later. It is absent when no later screen exists.

Write three fields in the first person and the present tense, as you would have written them at that moment, when you could see only the action history, the input state and the current game screen. The definitions below replace the ones in `Response Format`:

- `thought` (1-2 sentences): begin with what you see (the relevant objects, where they are relative to each other, the HUD values, the inputs currently held), then say why that calls for this action.
- `goal` (one short imperative phrase): the immediate objective this action serves. When the action history shows the goal of the previous step and you are still pursuing the same objective, reuse its wording; change the wording only when the objective changes.
- `expected_outcome` (1 sentence): the immediate visible result you expect from this action.

Rules:

- State as fact only what you could know at the moment of the decision. Whatever only the later screen reveals, such as the content of a hidden cell, a newly spawned tile or an obstacle that was not yet on screen, must not appear in any field.
- Use the later screen only to keep `expected_outcome` realistic: describe the result that was predictable in advance, and leave out details that chance decided. Phrase it as an expectation, never as an observation.
- If the action had no visible effect or went wrong, still write the result you were aiming for when you took it. Do not predict the failure.
- An empty action `{}` leaves every input unchanged. If something is held, you are continuing that input; if nothing is held, you are waiting. Explain why this is not yet the moment to change anything.
- Refer to things by what they are and where they are on the screen. Do not write pixel coordinates, JSON or channel names such as `keyboard.hold`.
- Write as the player in the middle of the game. Never mention this review, a recording, an annotation or the later screen.

Reply with exactly one JSON object with the keys `thought`, `goal` and `expected_outcome` (plus a one-sentence `reason` when the executed action is `{"complete": true}`), and nothing else. Do not repeat, add or alter device channels.
\end{lstlisting}
\end{revblock}

\subsection{Annotation Request}\label{sec:annotation-request}

The user message first presents the context of step $t$ in the same layout as a decision request under the game prompt.
It contains up to four preceding steps in chronological order, each with its screenshot and recorded action, followed by the input device state and the screenshot of step $t$.
The history shows only the recorded actions and none of the annotations of earlier steps, so that all steps of a trajectory are annotated independently and in parallel.
The action history is omitted at the first step of a trajectory.
For some games, the recorder stores the screenshot of a waiting step only at a fixed interval of steps, whereas every active step keeps its screenshot.
A screenshot that is not stored is replaced by the text ``(screenshot unavailable)'', and steps without a stored screenshot do not form training samples.

\begin{lstlisting}[style=prompt]
## Action History

Here are your last {n} steps, oldest first. For each step you can see the game screen you were looking at and the action you took for it.

### Step {j}

Game screen:
[screenshot of step j]
Action: {recorded action of step j}

...(one entry per preceding step)

## Current Input State

This is what is physically held right now, as confirmed by the environment. Base your `hold` / `release` decisions on it: held inputs stay down until you release them.
{input_state}

## Current Game Screen

This is the game screen right now. Decide your next action based on it, the current input state and your action history.
[screenshot of step t]
\end{lstlisting}

The executed action is then supplied using the following template, where \texttt{\{action\}} is the recorded action of step $t$ in JSON form.

\begin{revblock}
\begin{lstlisting}[style=prompt]
## Executed Action

This is the action you took on the current game screen above:
{action}
\end{lstlisting}
\end{revblock}

For a step that does not declare completion, the screenshot of the next step, i.e., the screen observed one step after the action, is appended with the following heading whenever it is stored, so that the expected outcome can reflect the actual effect of the action.

\begin{revblock}
\begin{lstlisting}[style=prompt]
## Game Screen After The Action

This is the game screen one step after you took that action:
[screenshot of step t+1]
\end{lstlisting}
\end{revblock}

A completion step, whose recorded action is \texttt{\{"complete": true\}}, has no subsequent screenshot.
Instead, the following note is appended to the executed action to request an additional justification.

\begin{revblock}
\begin{lstlisting}[style=prompt]
With this action you declared the task complete. Also include a one-sentence `reason` that says why the objective is satisfied on the current game screen.
\end{lstlisting}
\end{revblock}

The request ends with the following output instruction.
For a completion step, the JSON object in this instruction additionally contains the key \texttt{"reason"}.

\begin{revblock}
\begin{lstlisting}[style=prompt]
## Your Reply

Write down what was in your mind when you took the executed action. Reply with exactly one JSON object and no other text: {"thought": "...", "goal": "...", "expected_outcome": "..."}
\end{lstlisting}
\end{revblock}

We parse the reply as a JSON object and accept it when it contains nonempty \texttt{goal} and \texttt{thought} fields, together with a nonempty \texttt{reason} field for a completion step.
Otherwise, the request is repeated, up to five attempts in total.
If none of the attempts is accepted, the fields of the step are left empty, and the step is masked out of the training loss.
The accepted \texttt{thought} and \texttt{goal} fields form the \texttt{<think>} and \texttt{<goal>} segments of the training target in Appendix~\ref{sec:agent-response-format}, respectively, whereas the \texttt{expected\_outcome} field is not part of the training target.

\section{Implementation Details}\label{sec:implementation-details}

We leverage \tool to train Qwen3.5-4B, using LlamaFactory for action-focused SFT and verl for expert-guided on-policy learning.
We freeze the vision encoder and fine-tune the multimodal projector and language backbone.
We set the history window length to $K=\todo{1}$.
Action-focused SFT trains for \todo{3} epochs with an effective batch size of \todo{112}, a cosine-decayed learning rate of \todo{$5\times10^{-5}$}, a temporal decay rate of $\gamma=\todo{0.8}$, and a weight floor of $w_{\min}=\todo{0.2}$.
Initialized from the SFT policy, expert-guided on-policy learning rolls out $E=\todo{20}$ episodes per task at a cosine-decayed learning rate of \todo{$5\times10^{-6}$}, advancing each episode in truncated rollouts of $L=\todo{64}$ steps per policy update and sampling $G=\todo{8}$ candidate actions per step for ERPO.
For step-criticality weighting (Eq.~\eqref{eq:opi-weight}), we set the deviation horizon to $N=\todo{8}$ and the weighting strength to $\beta=\todo{3}$.
At inference, the agent uses greedy decoding and stops at \texttt{</act>}.
All experiments run on \todo{8} NVIDIA H200 GPUs.

\end{document}